\documentclass[10pt,twocolumn,letterpaper]{article}
\usepackage{cvpr}
\usepackage{times}
\usepackage{epsfig}
\usepackage{graphicx}
\usepackage{amsmath}
\usepackage{amssymb}
\usepackage{booktabs}
\usepackage{multirow}
\usepackage{enumerate}

\usepackage[utf8]{inputenc}
\usepackage{color}
\usepackage{graphicx,subcaption}
\usepackage{siunitx}
\usepackage{tabularx}
\usepackage{booktabs}
\usepackage{eso-pic}
\usepackage{lscape}
\RequirePackage{color}
\RequirePackage{graphicx,subcaption}
\RequirePackage{xspace}
\RequirePackage{eso-pic}

\usepackage{siunitx}
\usepackage{tabularx}
\usepackage{booktabs}
\usepackage{amssymb}
\usepackage{pifont} 
\usepackage{graphbox} 
\usepackage[font=small]{caption}

\usepackage{titling}
\pretitle{\begin{center}\Large \bf}
\posttitle{\par\vspace*{12pt}\end{center}}
\preauthor{\begin{center}\large \lineskip .5em \begin{tabular}[t]{c}}
\postauthor{\vspace*{1pt}\\\end{tabular}\par\end{center}}
\predate{}
\date{}
\postdate{}
\usepackage[pagebackref=true,breaklinks=true,letterpaper=true,colorlinks,bookmarks=false]{hyperref}

\cvprfinalcopy 

\def\cvprPaperID{5674} 

\ifcvprfinal\fi

\newcommand{\rot}[1]{\rotatebox{90}{#1}}

\makeatletter
\let\origparagraph\paragraph
\renewcommand\paragraph{\@ifstar{\starparagraph}{\nostarparagraph}}
\newcommand\nostarparagraph[1]
{\origparagraph{#1}\paragraphpostlude}
\newcommand\starparagraph[1]
{\paragraphprelude\origparagraph*{#1}\paragraphpostlude}
\newcommand\paragraphprelude{%
  \vspace{-13pt}
}
\newcommand\paragraphpostlude{%
}
\makeatother

\definecolor{mycol}{rgb}{0,.1,.4}

\usepackage[resetlabels]{multibib}
\newcites{Paper}{References}
\newcites{Suppl}{References}

\begin{document}

\title{Asynchronous Convolutional Networks for Object Detection \\ in Neuromorphic Cameras}

\author{Marco Cannici \hspace{20pt} 
	Marco Ciccone \hspace{20pt} 
	Andrea Romanoni\hspace{20pt} 
	Matteo Matteucci\\
	Politecnico di Milano, Italy\\
	{\tt\small \{marco.cannici,marco.ciccone,andrea.romanoni,matteo.matteucci\}@polimi.it}
}

\maketitle

\begin{abstract}
	Event-based cameras, also known as neuromorphic cameras, are bioinspired sensors able to perceive changes in the scene at high frequency with low power consumption.
	Becoming available only very recently, a limited amount of work addresses object detection on these devices.
	In this paper we propose two neural networks architectures for object detection:
	YOLE, which integrates the events into surfaces and uses a frame-based model to process them, and fcYOLE, an asynchronous event-based fully convolutional network which uses a novel and general formalization of the convolutional and max pooling layers to exploit the sparsity of camera events.
	We evaluate the algorithm with different extensions of publicly available datasets and on a novel synthetic dataset.
\end{abstract}

\section{Introduction}
\label{sec:intro}
Fundamental techniques underlying Computer Vision are based on the ability to extract meaningful features.
To this extent, Convolutional Neural Networks (CNNs) rapidly became the first choice in many computer vision applications such as image classification~\citePaper{Krizhevsky2012Dec, Simonyan2014, He2015Dec, Szegedy2016Feb}, object detection~\citePaper{Ren2015Jun, Redmon2015Jun, liu2016ssd}, semantic scene labeling~\citePaper{Yu2015Nov, Raj2015, Long2014Nov}, and they have been recently extended also to non-euclidean domains such as manifolds and graphs~\citePaper{kipf2017semi, monti2017geometric}. In most of the cases the input of these networks are images.

In the meanwhile, neuromorphic cameras~\citePaper{Serrano-Gotarredona2013Mar, Posch2011Jan, Berner2013} are becoming more and more widespread. These devices are bio-inspired vision sensors that attempt to emulate the functioning of biological retinas. As opposed to conventional cameras, which generate frames at a constant frame rate, these sensors output data only when a brightness change is detected in the field of view. Whenever this happens, an event $\mathbf{e} = \langle\, x, y, ts, p \,\rangle$ is generated indicating the position $(x, y)$, the instant $ts$ at which the change has been detected and its polarity $p \in \left\{1, -1\right\}$, \ie, if the brightness change is positive or negative. The result is a sensor able to produce a stream of asynchronous events that sparsely encodes changes with microseconds resolution and with minimum requirements in terms of power consumption and bandwidth.
The growth in popularity of these type of sensors, and their advantages in terms of temporal resolution and reduced data redundancy, have led to fully exploit the advantages of event-based vision for a variety of applications, \eg, object tracking~\citePaper{rameshlong, mitrokhin2018event, gehrig2018asynchronous}, visual odometry~\citePaper{Mueggler2016Oct,rebecq2017evo}, and optical flow estimation~\citePaper{bardow2016simultaneous,liu2018adaptive,stoffregen2018simultaneous}.

Spiking Neural Networks (SNNs)~\citePaper{Maass1997networks}, a processing model aiming to improve the biological realism of artificial neural networks, are one of the most popular neural model able to directly handle events. Despite their advantages in terms of speed and power consumption, however, training deep SNNs models on complex tasks is usually very difficult. To overcome the lack of scalable training procedures, recent works have focused on converting pre-trained deep networks to SNNs, achieving promising results even on complex tasks \citePaper{kim2019spiking, cao2015spiking, dieh2015spiking}.


An alternative solution to deal with event-based cameras is to make use of frame integration procedures and conventional frame-based networks~\citePaper{Perez-Carrasco2013Nov} which can instead rely on optimized training procedures. Recently, other alternatives to SNNs making use of hierarchical time surfaces \citePaper{lagorce2017hots} and memory cells \citePaper{sironi2018hats} have also been introduced. Another solution, proposed in \citePaper{Neil2016Oct}, suggests instead the use of LSTM cells to accumulate events and perform classification. An extension of this work making use of attention mechanisms has also been proposed in \citePaper{Cannici2019}.

Although event-cameras are becoming increasingly popular, only very few datasets for object detection in event streams are available, and a limited number of object detection algorithms has been proposed~\citePaper{Li2017, Chen2017Sep, Ramesh2017Oct}.

In this paper we introduce a novel hybrid approach to extract features for object detection problems using neuromorphic cameras. The proposed framework allows the design of object detection networks able to sparsely compute features while still preserving the advantages of conventional neural networks. More importantly, networks implemented using the proposed procedure are asynchronous, meaning that computation is only performed when a sequence of events arrive and only where previous results need to be recomputed. 

In Section \ref{sec:eventlayers} the convolution and max-pooling operations are reformulated by adding an internal state, \ie, a memory of the previous prediction, that allows us to sparsely recompute feature maps.
An asynchronous fully-convolutional network for event-based object detection which exploits this formulation is finally described in Section \ref{sec:fcyole}.


\section{Background}
\label{sec:basicModelYOLO}

\begin{figure*}
	\centering
	\includegraphics[width=0.98\textwidth]{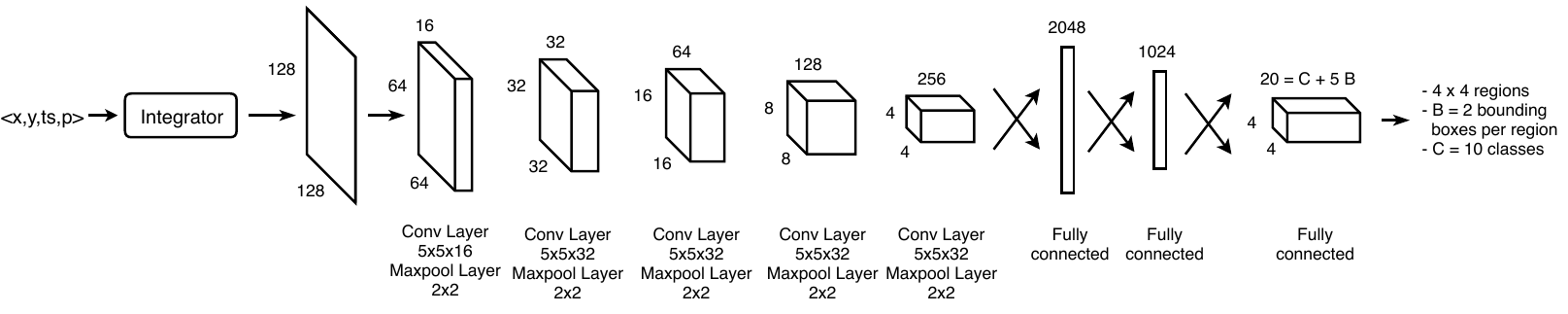}
	\vspace{-8pt}
	\caption{The YOLE detection network based on YOLO used to detect MNIST-DVS digits. The input surfaces are divided into a grid of $4 \times 4$ regions which predict $2$ bounding boxes each.}
	\label{fig:yolo_detection_x2}
\end{figure*}

\paragraph{Leaky Surface.}
The basic component of the proposed architectures is a procedure able to accumulate events. Sparse events generated by the neuromorphic camera are integrated into a \emph{leaky surface}, a structure that takes inspiration from the functioning of Spiking Neural Networks (SNNs) to maintain memory of past events. A similar mechanism has already been proposed in \citePaper{Cohen2016Sep}.
Every time an event with coordinates $(x_e, y_e)$ and timestamp $ts^t$ is received, the corresponding pixel of the surface is incremented of a fixed amount $\Delta_{incr}$.
At the same time, the whole surface is decremented by a quantity which depends on the time elapsed between the last received event and the previous one. 
The described procedure can be formalized by the following equations:
\begin{align}
\label{eq:integrator}
q_{x_s, y_s}^t &= max(p_{x_s, y_s}^{t-1} - \lambda \cdot \Delta_{ts}, 0) \\
p_{x_s, y_s}^t&=
\begin{cases}
q_{x_s, y_s}^t + \Delta_{incr} 	&if (x_s, y_s)^t = (x_e, y_e)^t\\
q_{x_s, y_s}^t 					&otherwise
\end{cases},
\end{align}
where  $p_{x_s, y_s}^t$ is the value of the surface pixel in position $(x_s, y_s)$ of the leaky surface and $\Delta_{ts}=ts^t - ts^{t-1}$. To improve readability in following equations, we name the quantity $(ts_{t} - ts_{t-1}) \cdot \lambda$ as $\Delta_{leak}$. Notice that the effects of $\lambda$ and $\Delta_{incr}$ are related: $\Delta_{incr}$ determines how much information is contained in each single event whereas $\lambda$ defines the decay rate of activations.
Given a certain choice of these parameters, similar results can be obtained by using, for instance, a higher increment $\Delta_{incr}$ and a higher temporal $\lambda$. For this reason, we fix $\Delta_{incr}=1$ and we vary only $\lambda$ based on the dataset to be processed. 
Pixel values are prevented from becoming negative by means of the max operation.

Other frame integration procedures, such as the one in \citePaper{Perez-Carrasco2013Nov}, divide the time in predefined constant intervals. Frames are obtained by setting each pixel to a binary value (depending on the polarity) if at least an event has been received in each pixel within the integration interval. With this mechanism however, time resolution is lost and the same importance is given to each event, even if it represents noise. The adopted method, instead, performs continuous and incremental integration and is able to better handle noise.

Similar procedures capable of maintaining time resolution have also been proposed, such as those that make use of exponential decays \citePaper{Cohen2016Sep, Lagorce2016Jul} to update surfaces, and those relying on histograms of events \citePaper{Maqueda_2018_CVPR}. Recently, the concept of \emph{time surface} has also been introduced in \citePaper{lagorce2017hots} where surfaces are obtained by associating each event with temporal features computed applying exponential kernels to the event neighborhood. Extensions of this procedure making use of memory cells \citePaper{sironi2018hats} and event histograms \citePaper{AlexZihaoZhu_2018} have also been proposed. Although these event representations better describe complex scene dynamics, we make use of a simpler formulation to derive a linear dependence between consecutive surfaces. This allows us to design the event-based framework discussed in Section \ref{sec:eventlayers} in which time decay is applied to every layer of the network.

\paragraph*{Event-based Object Detection.}
We identified YOLO \citePaper{Redmon2015Jun} as a good candidate model to tackle the object detection problem in event-based scenarios: it is fully-differentiable and produces predictions with small input-output delays.
By means of a standard CNN and with a single forward pass, YOLO is able to simultaneously predict not only the class, but also the position and dimension of every object in the scene.
We used the YOLO loss and the previous leaky surface procedure to train a baseline model which we called YOLE: "You Only Look at Events". The architecture is depicted in Figure \ref{fig:yolo_detection_x2}. We use this model as a reference to highlight the strengths and weaknesses of the framework described in Section \ref{sec:eventlayers}, which is the main contribution of this work.
YOLE processes $128 \times 128$ surfaces, it predicts $B = 2$ bounding boxes for each region and classifies objects into $C$ different categories. 

Note that in this context, we use the term YOLO to refer only to the training procedure proposed by \citePaper{Redmon2015Jun} and not to the specific network architecture. We used indeed a simpler structure for our models as explained in Section \ref{sec:exp}.
Nevertheless, YOLE, \ie, YOLO + leaky surface, does not exploit the sparse nature of events; to address this issue, in the next section, we propose a fully event-based asynchronous framework for convolutional networks.

\section{Event-based Fully Convolutional Networks}
\label{sec:eventlayers}
Conventional CNNs for video analysis treat every frame independently and recompute all the feature maps entirely, even if consecutive frames differ from each other only in small portions. 
Beside being a significant waste of power and computations, this approach does not match the nature of event-based cameras.

To exploit the event-based nature of neuromorphic vision, we propose a modification of the forward pass of fully convolutional architectures. In the following the convolution and pooling operations are reformulated to produce the final prediction by recomputing only the features corresponding to regions affected by the events.
Feature maps maintain their state over time and are updated only as a consequence of incoming events. At the same time, the leaking mechanism that allows past information to be forgotten, acts independently on each layer of the CNN. This enables features computed in the past to fade away as their visual information starts to disappear in the input surface.
The result is an asynchronous CNN able to perform computation only when requested and at different rates. The network can indeed be used to produce an output only when new events arrive, dynamically adapting to the timings of the input, or to produce results at regular rates by using the leaking mechanism to update layers in absence of new events.

The proposed framework has been developed to extend the YOLE detection network presented in Section~\ref{sec:basicModelYOLO}. Nevertheless, this method can be applied to any convolutional architecture to perform asynchronous computation. A CNN trained to process frames reconstructed from streams of events can indeed be easily converted into an event-based CNN without any modification on its layers composition, and by using the same weights learned while observing frames, maintaining its output unchanged.

\subsection{Leaky Surface Layer}
The procedure used to compute the leaky surface described in Section \ref{sec:basicModelYOLO} is embedded into an actual layer of the proposed framework.
Furthermore, to allow subsequent layers to locate changes inside the surface, the following information are also forwarded to the next layer:
(i) the list of incoming events.
(ii) $\Delta_{leak}$, which is sent to all the subsequent layers to update feature maps not affected by the events.
(iii) the list of surface pixels which have been reset to $0$ by the max operator in Equation \eqref{eq:integrator}.

\subsection{Event-based Convolutional Layer (e-conv)}
\label{sec:eventBasedConv}

The \emph{event-based convolutional} (e-conv) layer we propose uses events to determine where the input feature map has changed with respect to the previous time step and, therefore, which parts of its \emph{internal state}, \ie, the feature map computed at the previous time step, must be recomputed and which parts can be reused. 
We use a procedure similar to the one described in the previous section to let features decay over time. However, due to the transformations applied by previous layers and the composition of their activation functions, $\Delta_{leak}$ may act differently in different parts of the feature map. 
For instance, the decrease of a pixel intensity value in the input surface may cause the value computed by a certain feature in a deeper layer to decrease, but it could also cause another feature of the same layer to increase. The update procedure, therefore, must also be able to accurately determine how a single bit of information is transformed by the network through all the previous layers, in any spatial location. 
We face this issue by storing an additional feature map, $F_{(n)}$, and by using a particular class of activation functions in the hidden layers of the network. 

Let's consider the first layer of a CNN which processes surfaces obtained using the procedure described in the previous section and which computes the convolution of a set of filters $W$ with bias $b$ and activation function $g(\cdot)$. The computation performed on each receptive field is:
\begin{equation}
\label{eq:rf_conv_t0}
y_{ij_{(1)}}^t = g\left( \sum_h \sum_k x_{h+i,k+j}^t W_{hk_{(1)}} \,\,+ b_{_{(1)}}\right) = g(\tilde{y}_{ij_{(1)}}^t),
\end{equation}
where $h, k$ select a pixel $x_{h+i,k+j}^t$ in the receptive field of the output feature $(i,j)$ and its corresponding value in the kernel $W$, whereas the subscript $(1)$ indicates the hidden-layer of the network (in this case the first after the leaky surface layer).

When a new event arrives, the leaky surface layer decreases all the pixels by $\Delta_{leak}$, \ie, a pixel not directly affected by the event becomes: $x_{hk}^{t+1} = x_{hk}^t - \Delta_{leak}^{t+1}$, with $\Delta_{leak}^{t+1} > 0$. At time $t+1$ Equation \eqref{eq:rf_conv_t0} becomes:

{
	\begin{equation}
	\label{eq:rf_conv_t1}
	\begin{split}
	y_{ij_{(1)}}^{t+1} 	 & = g\left( \sum_h \sum_k x_{h+i,k+j}^{t+1} W_{hk_{(1)}} \,\,+ b_{_{(1)}}\right) \\
	& = g\left( \sum_h \sum_k (x_{h+i,k+j}^t - \Delta_{leak}^{t+1}) W_{hk_{(1)}} \,\,+ b_{_{(1)}}\right) \\
	& = g\Biggl( \tilde{y}_{ij_{(1)}}^t - \Delta_{leak}^{t+1} \sum_h \sum_k W_{hk_{(1)}} \Biggr).
	\end{split}
	\end{equation}
	
}
If $g(\cdot)$ is (i) a piecewise linear activation function $g(x) = \{ \alpha_i \cdot x \quad \text{if $x \in D_i$}\}$, as ReLU or Leaky ReLU, and we assume that (ii) the updated value does not change which linear segment of the activation function the output falls onto and, in this first approximation, (iii) the leaky surface layer does not restrict pixels using $\max(\cdot, 0)$, Equation \ref{eq:rf_conv_t1} can be rewritten as it follows:
\begin{equation}
\label{eq:rf_conv_t1_actfn}
y_{ij_{(1)}}^{t+1} = y_{ij_{(1)}}^t - \Delta_{leak}^{t+1} \alpha_{ij_{(1)}} \sum_h \sum_k W_{hk_{(1)}},
\end{equation}
where $\alpha_{ij_{(1)}}$ is the coefficient applied by the piecewise function $g(\cdot)$ which depends on the feature value at position $(i, j)$. 
When the previous assumption is not satisfied, the feature is recomputed as its receptive field was affected by an event (\ie, applying the filter $W$ locally to $x^{t+1}$).

\begin{figure*}[t]
	\centering
	\includegraphics[width=0.95\textwidth]{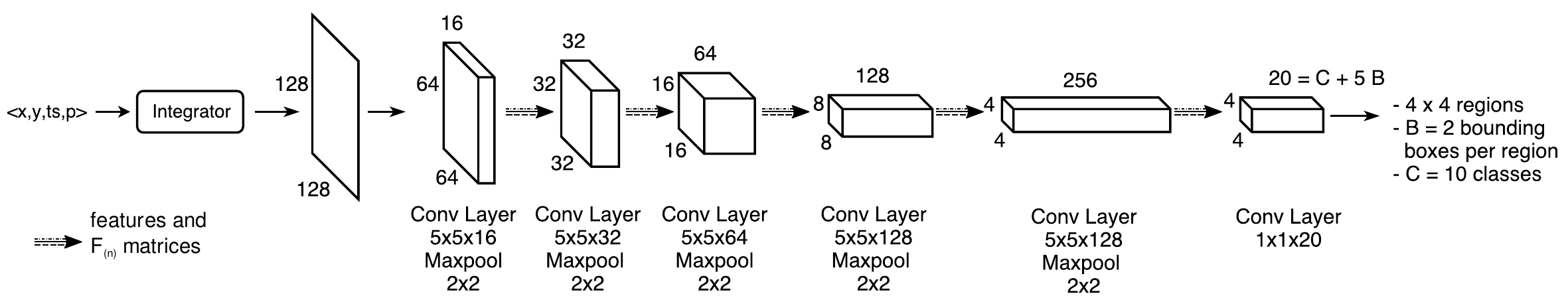}
	\caption{fcYOLE: a fully-convolutional detection network based on YOLE. The last layer is used to map the feature vectors into a set of $20$ values which define the parameters of the predicted bounding boxes.}
	\label{fig:sim_yolo_net}
\end{figure*}

Consider now a second convolutional layer attached to the first one:
\begin{equation}
\label{eq:rf_conv_layer2}{
	\tiny
	\begin{split}
	&y_{ij_{(2)}}^{t+1} 		= g\left( \sum_{h,k} y_{i+h,j+k_{(1)}}^{t+1} W_{hk_{(2)}} \,\,+ b_{_{(2)}}\right) \\
	& = g\left( \sum_{h,k} \left( y_{i+h,j+k_{(1)}}^t - \Delta_{leak}^{t+1} \alpha_{i+h,j+k_{(1)}} \sum_{h',k'} W_{h'k'_{(1)}} \right) W_{hk_{(2)}} + b_{_{(2)}}\right) \\
	& = y_{ij_{(2)}}^t - \Delta_{leak}^{t+1} \alpha_{ij_{(2)}} \sum_{h,k} \left( \alpha_{i+h,j+k_{(1)}} \sum_{h',k'} W_{h'k'_{(1)}} \right)  W_{hk_{(2)}} \\
	& = y_{ij_{(2)}}^t - \Delta_{leak}^{t+1} \alpha_{ij_{(2)}} \sum_{h,k} \mathit{F}^{t+1}_{h+i,k+j_{(1)}} W_{hk_{(2)}} = y_{ij_{(2)}}^t - \Delta_{leak}^{t+1} \mathit{F}^{t+1}_{ij_{(2)}}.
	\end{split}
}
\end{equation}
The previous equation can be extended by induction as it follows:
\begin{equation}
\label{eq:actfn_matrix}
\begin{gathered}
y_{ij_{(n)}}^{t+1} 	= y_{ij_{(n)}}^t - \Delta_{leak}^{t+1} \mathit{F}^{t+1}_{ij_{(n)}}, \\ \text{with}\,\, 
\mathit{F}^{t+1}_{ij_{(n)}} = \alpha_{ij_{(n)}} \sum_h \sum_k \mathit{F}^{t+1}_{i+h,j+k_{(n-1)}} W_{hk_{(n)}} \,\, \text{if} \,\, n > 1 \, ,
\end{gathered}
\end{equation}
where $\mathit{F}_{ij_{(n)}}$ expresses how visual inputs are transformed by the network in every receptive field $(i,j)$, \ie, the composition of the previous layers activation functions.

Given this formulation, the $\max$ operator applied by the leaky surface layer can be interpreted as a ReLU, and Equation \eqref{eq:rf_conv_t1_actfn} becomes:

\begin{equation}
\label{eq:rf_conv_layer1}
y_{ij_{(1)}}^{t+1} = y_{ij_{(1)}}^t - \Delta_{leak}^{t+1} \alpha_{ij_{(1)}} \sum_h \sum_k \mathit{F}^{t+1}_{i+h,j+k_{(0)}} W_{hk_{(1)}} ,
\end{equation}
where the value $\mathit{F}_{i+h,j+k_{(0)}}$ is $0$ if the pixel $x_{i+h,j+k} \leq 0$ and $1$ otherwise.

Notice that $\mathit{F}_{ij_{(n)}}$ needs to be updated only when the corresponding feature changes enough to make the activation function use a different coefficient $\alpha$, \eg, from 0 to 1 in case of ReLU. In this case $\mathit{F}_{(n)}$ is updated locally in correspondence of the change by using the update matrix of the previous layer and by applying Equation \ref{eq:actfn_matrix} only for the features whose activation function has changed. 
Events are used to communicate the change to subsequent layers so that their update matrix can also be updated accordingly. 

\begin{figure*}
	\centering
	\begin{subfigure}[c]{0.5\textwidth}
		\begin{center}
			\centering
			\includegraphics[height=0.4\textwidth]{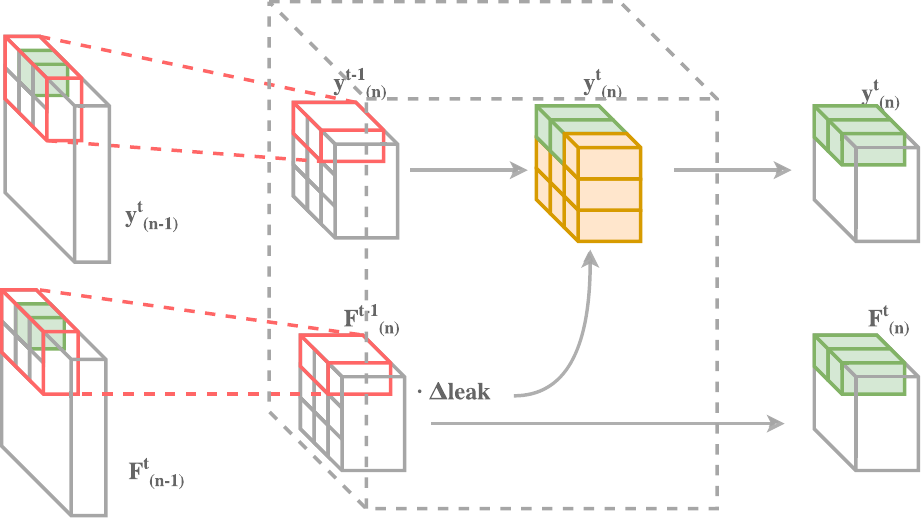}
			\vspace{-0.1cm}
			\caption{} \label{fig:event-based_conv}   
		\end{center}
	\end{subfigure}%
	\hfill
	\begin{subfigure}[c]{0.5\textwidth}
		\centering
		\includegraphics[height=0.4\textwidth]{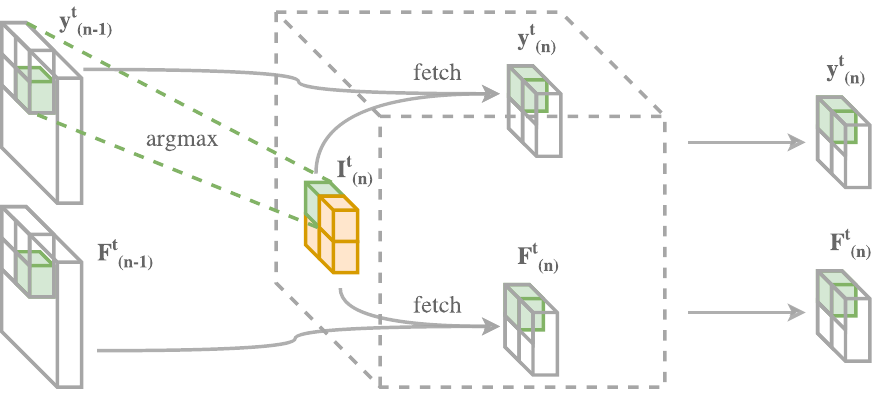}
		\vspace{-0.1cm}
		\caption{} \label{fig:event-based_maxpool}
	\end{subfigure}%
	
	\vspace{-0.35cm}
	\caption{
		The structure of the e-conv \textbf{(a)} and e-max-pooling layers \textbf{(b)}. The internal states and the update matrices are recomputed locally only where events are received (green cells) whereas the remaining regions (depicted in yellow) are obtained reusing the previous state.}
\end{figure*}

The internal state of the e-conv layer, therefore, comprises the feature maps $y_{(n)}^{t-1}$ and the update values $\mathit{F}_{{(n)}}^{t-1}$ computed at the previous time step. 
The initial values of the internal state are computed making full inference on a blank surface; this is the only time the network needs to be executed entirely. 
As a new sequence of events arrives the following operations are performed (see Figure \ref{fig:event-based_conv}):
\begin{enumerate}[i.]
	\itemsep0pt 
	\item Update $\mathit{F}_{{(n)}}^{t-1}$ locally on the coordinates specified by the list of incoming events (Eq. \eqref{eq:actfn_matrix}). Note that we do not distinguish between actual events and those generated by the use of a different slope in the linear activation function.
	\item Update the feature map $y_{(n)}$ with Eq. \eqref{eq:actfn_matrix} in the locations which are not affected by any event and generate an output event where the activation function coefficient has changed.
	\item Recompute $y_{(n)}$ by applying $W$ locally in correspondence of the incoming events and output which receptive field has been affected.
	\item Forward the feature map and the events generated in the current step to the next layer.
\end{enumerate}




\subsection{Event-based Max Pooling Layer (e-max-pool)}
\label{sec:eventBasedPool}

The location of the maximum value in each receptive field of a max-pooling layer is likely to remain the same over time.
An event-based pooling layer, hence, can exploit this property  to avoid recomputing every time the position of maximum values. 

The internal state of an event-based max-pooling (e-max-pool) layer can be described by a \emph{positional matrix} $\mathit{I}_{(n)}^t$, which has the shape of the output feature map produced by the layer, and which stores, for every receptive field, the position of its maximum value. 
Every time a sequence of events arrives, the internal state $\mathit{I}_{(n)}^{t}$ is sparsely updated by recomputing the position of the maximum values in every receptive field affected by an incoming event.
The internal state is then used both to build the output feature map and to produce the \emph{update matrix} $\mathit{F}_{(n)}^{t}$ by fetching the previous layer on the locations provided by the indices $\mathit{I}_{ij_{(n)}}^t$.
For each e-max-pool layer, the indices of the receptive fields where the maximum value changes are communicated to the subsequent layers so that the internal states can be updated accordingly. This mechanism is depicted in Figure \ref{fig:event-based_maxpool}.

Notice that the leaking mechanism acts differently in distinct regions of the input space. Features inside the same receptive field can indeed decrease over time with different speeds as their update values $\mathit{F}_{ij_{(n)}}^{t}$ could be different. Therefore, even if no event has been detected inside a region, the position of its maximum value might change.
However, if an input feature $M$ has the minimum update rate $\mathit{F}_{M_{(n-1)}}$ among features in its receptive field $R$ and it also corresponds to the maximum value in $R$, the corresponding output feature will decrease slower than all the others in $R$ and its value will remain the maximum. In this case, its index $\mathit{I}_{(n)_R}^t$ does not need to be recomputed until a new event arrives in $R$. We check if the maximum has to be recomputed for each receptive field affected by incoming events and also in all positions where the previous condition does not hold.

\subsection{Event FCN for Object Detection (fcYOLE)}
\label{sec:fcyole}
To fully exploit the event-based layers presented so far, the YOLE model described in Section \ref{sec:basicModelYOLO} needs to be converted into a fully convolutional object detection network replacing all its layers with their event-based versions (see Figure \ref{sec:eventlayers}). Moreover, fully-connected layers are replaced with $1 \times 1$ e-conv layers which map features extracted by the previous layers into a precise set of values defining the bounding boxes parameters predicted by the network. Training was first performed on a network composed of standard layers; the learned weights were then used with e-conv and e-max-pool layers during inference.

This architecture divides the $128 \times 128$ field of view into a grid of $4 \times 4$ regions that predicts $2$ bounding boxes each and classify the detected objects into $C$ different classes. The last $1 \times 1$ e-conv layer is used to decrease the dimensionality of the feature vectors and to map them into the right set of parameters, regardless of their position in the field of view.

Moreover, this architecture can be used to process surfaces of different sizes without the need to re-train or re-design it. The subnetworks processing $32 \times 32$ regions, in fact, being defined by the same set of parameters, can be stacked together to process even larger surfaces.


\section{Experiments}
\label{sec:exp}

\subsection{Datasets}
Only few event-based object recognition datasets are publicly available in the literature. The most popular ones are: N-MNIST \citePaper{Orchard2015Nov}, MNIST-DVS \citePaper{Serrano-Gotarredona2015Dec}, CIFAR$10$-DVS \citePaper{Li2017May}, N-Caltech101 \citePaper{Orchard2015Nov} and POKER-DVS~\citePaper{Serrano-Gotarredona2015Dec}. These datasets are obtained from the original MNIST \citePaper{Lecun1998Nov}, CIFAR-$10$ \citePaper{Krizhevsky2012} and Caltech101 \citePaper{Fei-Fei2006Apr} datasets by recording the original images with an event camera while moving the camera itself or the images of the datasets.
We performed experiments on N-Caltech101 and on modified versions of N-MNIST and MNIST-DVS for object detection, \ie, \textit{Shifted N-MNIST} and \textit{Shifted MNIST-DVS}, and on an extended version of POKER-DVS, namely \textit{OD-Poker-DVS}.
Moreover we also perform experiments on a synthetic dataset, named \textit{Blackboard MNIST}, showing digits written on a blackboard. A detailed description of these datasets is provided in the supplementary materials.

\paragraph*{Shifted N-MNIST}
The N-MNIST \citePaper{Orchard2015Nov} dataset is a conversion of the popular MNIST \citePaper{Lecun1998Nov} image dataset for computer vision. We enhanced this collection by building a slightly more complex set of recordings. Each sample is indeed composed of two N-MNIST samples placed in two random non-overlapping locations of a bigger $124 \times 124$ field of view. Each digit was also preprocessed by extracting its bounding box which was then moved, along with the events, in its new position of the bigger field of view. The final dataset is composed of $60,000$ training and $10,000$ testing samples.

\paragraph*{Shifted MNIST-DVS}
We used a similar procedure to obtain Shifted MNIST-DVS recordings. We first extracted bounding boxes with the same procedure used in Shifted N-MNIST and then placed them in a $128 \times 128$ field of view. We mixed MNIST-DVS \emph{scale4}, \emph{scale8} and \emph{scale16} samples within the same recording obtaining a dataset composed of $30,000$ samples.

\paragraph*{OD-Poker-DVS}
The Poker-DVS dataset is a small collection of neuromorphic samples showing poker card pips obtained by extracting $31 \times 31$ symbols from three deck recordings. We used the tracking algorithm provided with the dataset to track pips and enhance the original uncut deck recordings with their bounding boxes. We finally divided these recordings into a set of shorter examples obtaining a collection composed of $218$ training and $74$ testing samples.

\paragraph*{Blackboard MNIST}
We used the DAVIS simulator released by \citePaper{Mueggler2016Oct} to build our own set of synthetic recordings. The resulting dataset consists of a number of samples showing digits written on a blackboard in random positions and with different scales. We preprocessed a subset of images from the original MNIST dataset by removing their background and by making them look as if they were written with a chalk. Sets of digits were then placed on the image of a blackboard and the simulator was finally run to obtain event-based recordings and the bounding boxes of every digit visible within the camera field of view. The resulting dataset is the union of three simulations featuring increasing level of variability in terms of camera trajectories and digit dimensions. The overall dataset is composed of $2750$ training and $250$ testing samples.

\paragraph*{N-Caltech101} The N-Caltech101 \citePaper{Orchard2015Nov} collection is the only publicly available event-based dataset providing bounding boxes annotations. We split the dataset into $80\%$ training and $20\%$ testing samples using a stratified split. Since no ground truth bounding boxes are available for the \textit{background} class, we decided not to use this additional category in our experiments.

\subsection{Experiments Setup}
Event-based datasets, especially those based on MNIST, are generally simpler than the image-based ones used to train the original YOLO architecture. Therefore, we designed the MNIST object detection networks taking inspiration from the simpler LeNet \citePaper{Lecun1998Nov} model composed of $6$ conv-pool layers for feature extraction. Both YOLE and fcYOLE share the same structure up to the last regression/classification layers.

For what concerns the N-Caltech101 dataset, we used a slightly different architecture inspired by the structure of the VGG16 model~\citePaper{Simonyan2014}. The network is composed by only one layer for each group of convolutional layers, as we noticed that a simpler architecture achieved better results. Moreover, the dimensions of the last fully-connected layers have been adjusted such that the surface is divided into a grid of $5 \times 7$ regions predicting $B = 2$ bounding boxes each. As in the original YOLO architecture we used Leaky ReLU for the activation functions of hidden layers and a linear activation for the last one.

In all the experiments the first $4$ convolutional layers have been initialized with kernels obtained from a recognition network pretrained to classify target objects, while the remaining layers using the procedure proposed in~\citePaper{Glorot2010Mar}. All networks were trained optimizing the multi-objective loss proposed by~\citePaper{Redmon2015Jun} using Adam~\citePaper{Kingma2014Dec}, learning rate $10^{-4}$, $\beta_1 = 0.9$, $\beta_2 = 0.999$ and $\epsilon=10^{-8}$. The batch size was chosen depending on the dataset: $10$ for Shifted N-MNIST, $40$ for Shifted MNIST-DVS and N-Caltech101, $25$ for Blackboard MNIST and $35$ for Poker-DVS with the aim of filling the memory of the GPU optimally. Early-stopping was applied to prevent overfitting using validation sets with the same size of the test set.

\subsection{Results and Discussion}

\begin{table*}[!htbp]
	\centering
	\scriptsize
	\caption{YOLE Top-20 average precisions on N-Caltech101. Full table provided in the supplemental material.}
	\vspace{-8pt}
	\begin{tabularx}{\textwidth}{c|*{20}{X}}
		& \rot{Motorbikes} & \rot{airplanes} & \rot{\parbox{1cm}{Faces\\easy}} & \rot{metronome} & \rot{laptop} & \rot{\parbox{1cm}{dollar\\bill}} & \rot{umbrella} & \rot{watch} & \rot{minaret} & \rot{\parbox{1cm}{grand\\piano}} & \rot{menorah} & \rot{\parbox{1cm}{inline\\skate}} & \rot{saxophone} & \rot{stapler} & \rot{\parbox{1cm}{windsor\\chair}} & \rot{rooster} & \rot{\parbox{1cm}{yin\\yang}} & \rot{Leopards} & \rot{trilobite} & \rot{garfield} \\
		\midrule 
		
		AP & 97.8 & 95.8 & 94.7 & 88.3 & 88.1 & 86.5 & 85.9 & 84.2 & 81.3 & 81.3 & 80.7 & 75.1 & 68.4 & 68.1 & 65.2 & 64.5 & 63.3 & 62.9 & 62.5 & 62.3 \\
		
		$N_{train}$ & 480 & 480 & 261 & 20 & 49 & 32 & 45 & 145 & 46 & 61 & 53 & 19 & 24 & 27 & 34 & 31 & 36 & 120 & 52 & 22 \\
	\end{tabularx}
	\label{tab:YOLE_ncaltech101}
	\vspace{-3pt}
\end{table*}
\begin{table*}[!htbp]
	\centering
	\scriptsize
	\caption{fcYOLE Top-20 average precisions on N-Caltech101. Full table provided in the supplemental material.}
	\vspace{-8pt}
	\begin{tabularx}{\textwidth}{c|*{20}{X}}
		& \rot{Motorbikes} & \rot{airplanes} & \rot{\parbox{1cm}{Faces\\easy}} & \rot{watch} & \rot{\parbox{1cm}{dollar\\bill}} & \rot{\parbox{1cm}{car\\side}} & \rot{\parbox{1cm}{grand\\piano}} & \rot{menorah} & \rot{metronome} & \rot{umbrella} & \rot{\parbox{1cm}{yin\\yang}} & \rot{saxophone} & \rot{minaret} & \rot{\parbox{1cm}{soccer\\ball}} & \rot{Leopards} & \rot{dragonfly} & \rot{\parbox{1cm}{stop\\sign}} & \rot{\parbox{1cm}{windsor\\chair}} & \rot{accordion} & \rot{buddha} \\
		\midrule 
		
		AP & 97.5 & 96.8 & 92.2 & 75.7 & 74.4 & 70.3 & 69.5 & 67.7 & 63.4 & 61.0 & 60.4 & 59.7 & 59.5 & 57.3 & 57.2 & 55.6 & 55.1 & 52.3 & 48.3 & 46.5 \\
		
		$N_{train}$ & 480 & 480 & 261 & 145 & 32 & 75 & 61 & 53 & 20 & 45 & 36 & 24 & 46 & 40 & 120 & 42 & 40 & 34 & 33 & 51 \\
	\end{tabularx}
	\label{tab:fcYOLE_ncaltech101}
\end{table*}

\begin{table}[t]
	\caption{Performance comparison between YOLE and fcYOLE.}
	\vspace{-8pt}
	\label{tab:results}
	\centering
	\small
	\begin{tabular}{cccccccc}
		
		\multicolumn{4}{c}{S-MNIST-DVS} & \multicolumn{4}{c}{Blackboard MNIST} \\
		\multicolumn{2}{c}{fcYOLE} & \multicolumn{2}{c}{YOLE} & \multicolumn{2}{c}{fcYOLE} & \multicolumn{2}{c}{YOLE} \\
		\hline
		acc & mAP & acc & mAP & acc & mAP & acc & mAP \\
		94.0 & 87.4 & 96.1 & 92.0 & 88.5 & 84.7 & 90.4 & 87.4  \vspace{5pt} \\
		
		\multicolumn{4}{c}{OD-Poker-DVS} & \multicolumn{4}{c}{N-Caltech101} \\
		\multicolumn{2}{c}{fcYOLE} & \multicolumn{2}{c}{YOLE} & \multicolumn{2}{c}{fcYOLE} & \multicolumn{2}{c}{YOLE} \\
		\hline
		acc & mAP & acc & mAP & acc & mAP & acc & mAP \\
		79.10 & 78.69 & 87.3 & 82.2 & 57.1 & 26.9 & 64.9 & 39.8 \\
		
	\end{tabular}
\end{table}

\begin{table}[t]
	\begin{center}
		\caption{YOLE performance on S-N-MNIST variants.}
		\vspace{-8pt}
		\label{tab:resNMNIST}
		\centering
		\small
		\setlength{\tabcolsep}{3px}
		\begin{tabular}{lccccc}
			&\multicolumn{5}{c}{S-N-MNIST} \\ 
			&v1&v2&v2*&v2fr&v2fr+ns \\ 
			\hline
			accuracy &94.9&91.7&94.7&88.6&85.5  \\
			mAP	  &91.3&87.9&90.5&81.5&77.4  \\
		\end{tabular}
	\end{center}
	\vspace{-0.8cm}
\end{table}

\begin{figure*}[tp]
	\centering
	\setlength{\tabcolsep}{1px}
	\begin{tabular}{ccccc}
		Shifted & Shifted &  & & Blackboard\\
		N-MNIST & MNIST-DVS & OD-Poker-DVS & N-Caltech101 & MNIST\\[0.05cm]
		\includegraphics[width=0.18\textwidth]{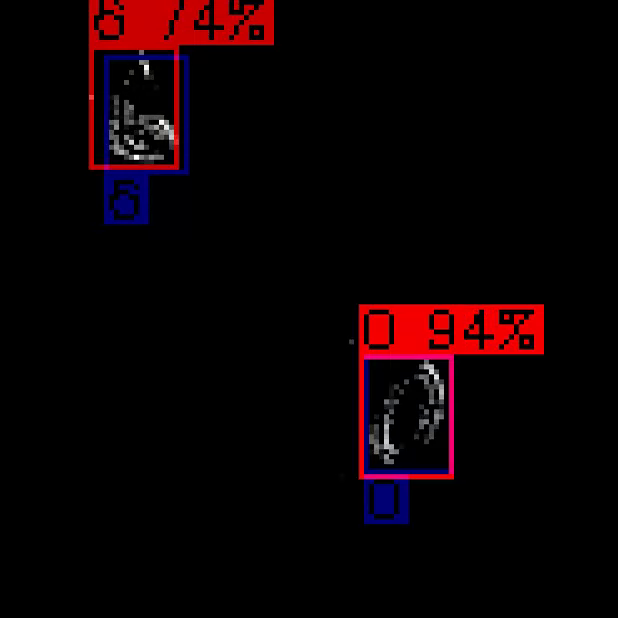}&
		\includegraphics[width=0.18\textwidth]{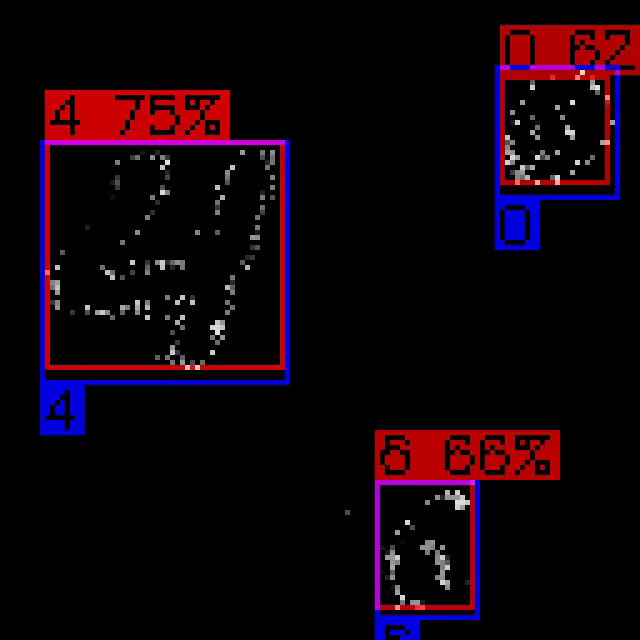}&
		\includegraphics[width=0.18\textwidth]{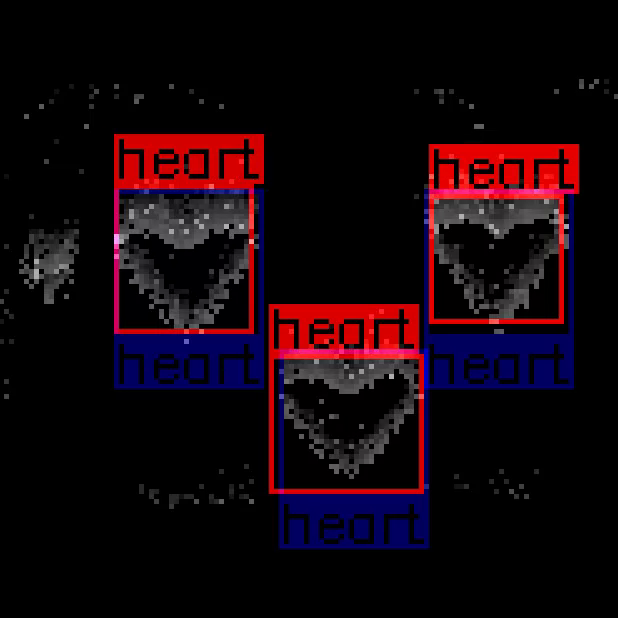}&
		\includegraphics[width=0.25\textwidth]{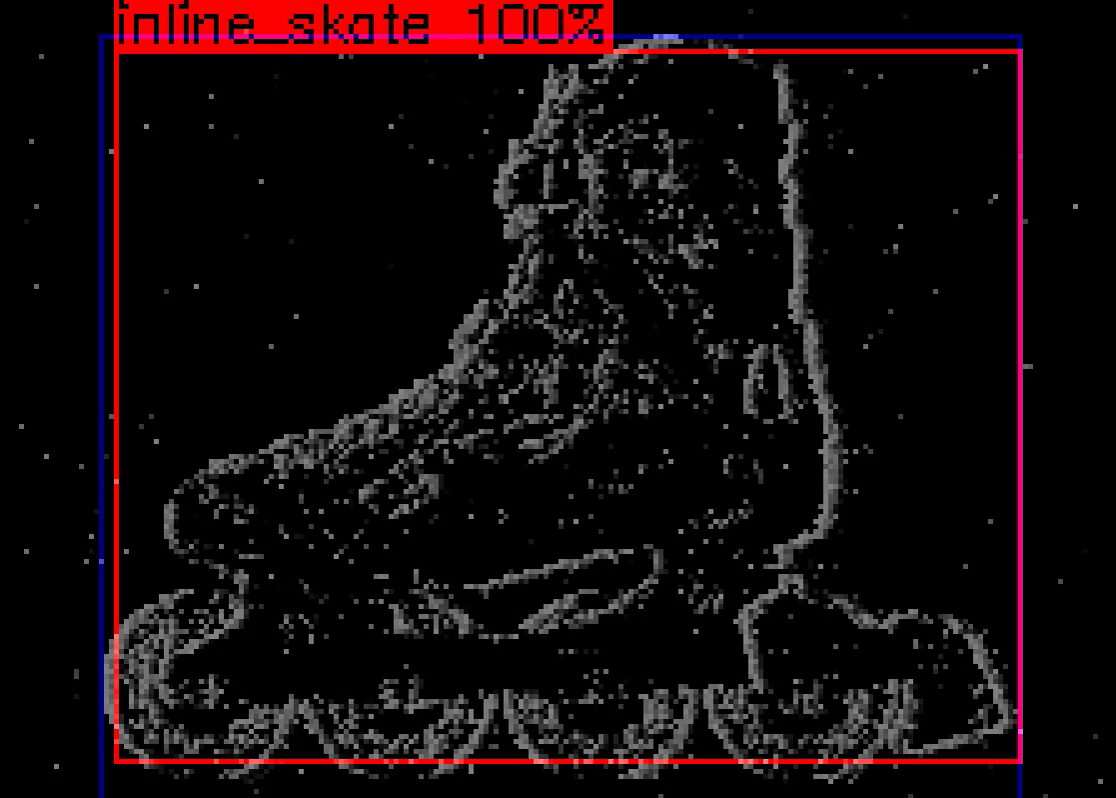}&
		\includegraphics[width=0.18\textwidth]{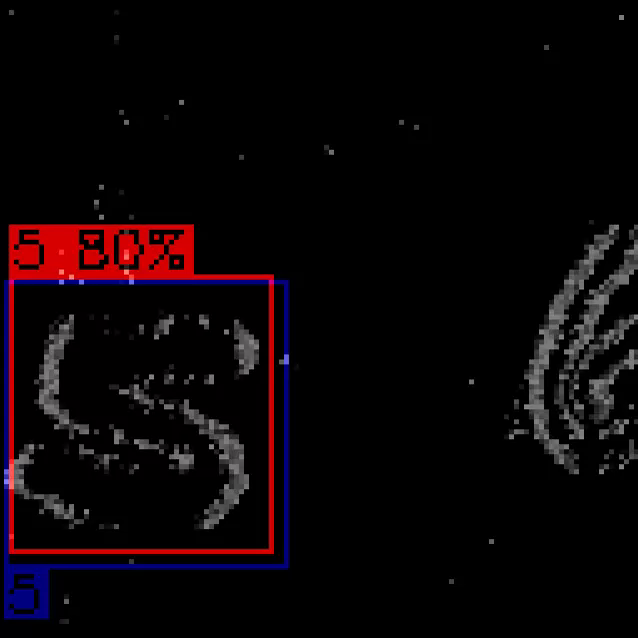}\\
		\includegraphics[width=0.18\textwidth]{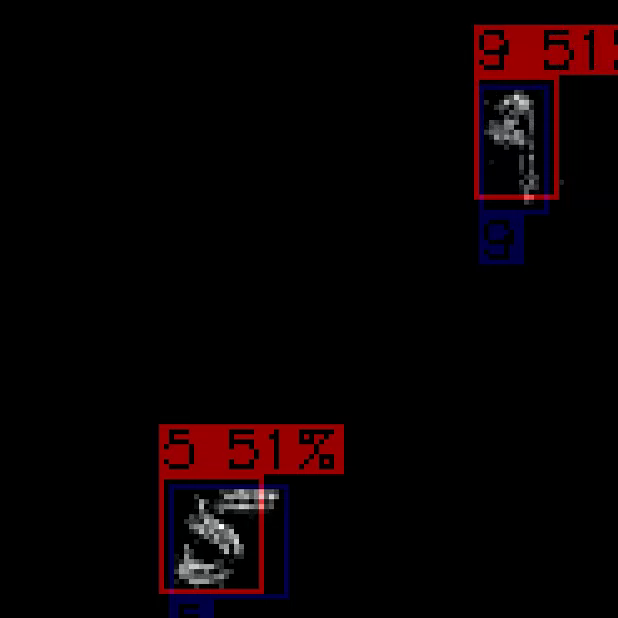}&
		\includegraphics[width=0.18\textwidth]{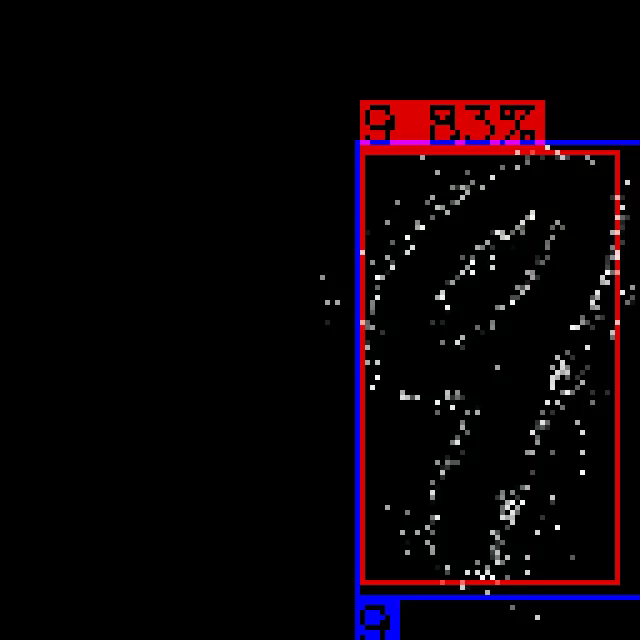}&
		\includegraphics[width=0.18\textwidth]{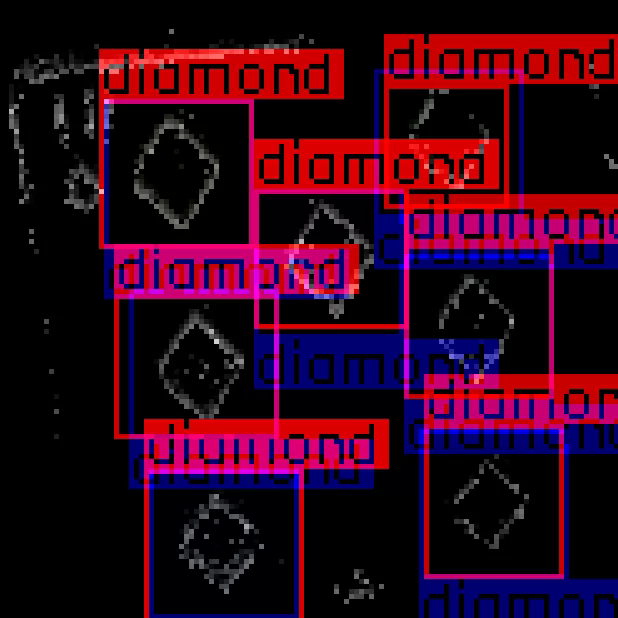}&
		\includegraphics[width=0.25\textwidth]{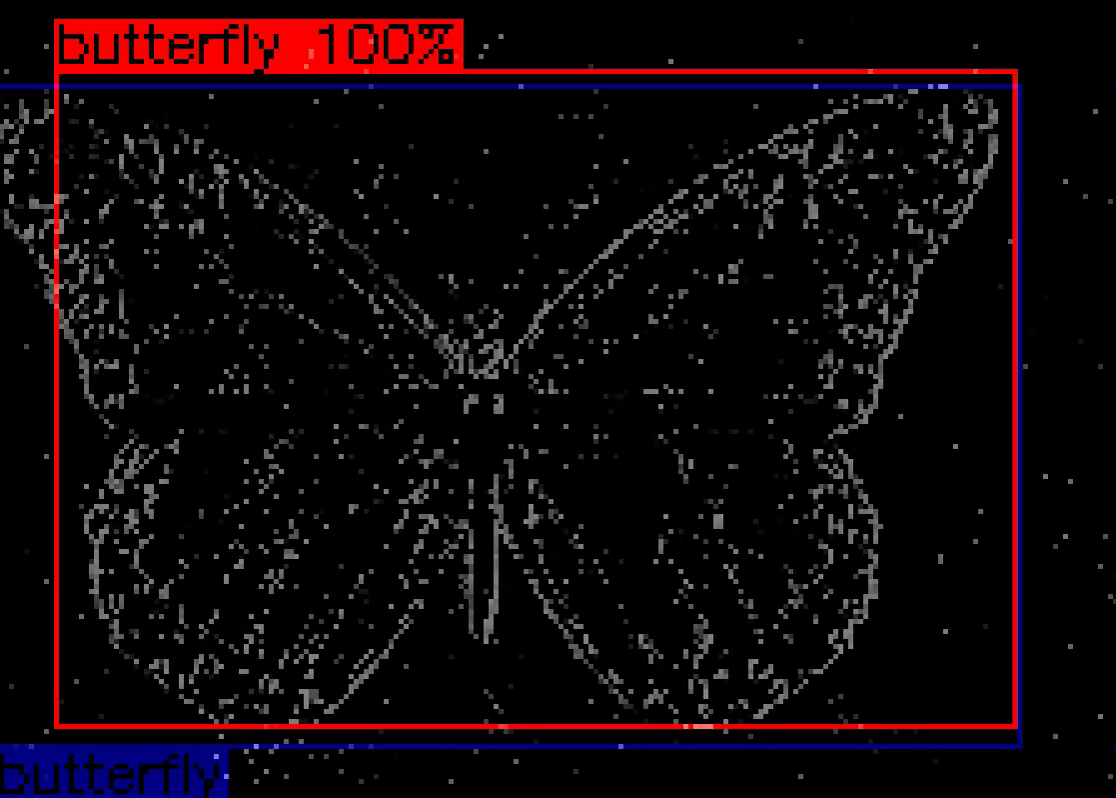}&
		\includegraphics[width=0.18\textwidth]{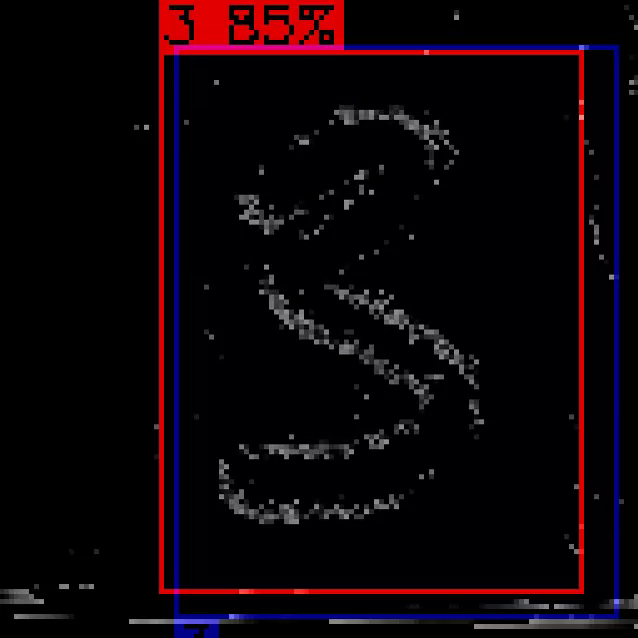}\\
		\includegraphics[width=0.18\textwidth]{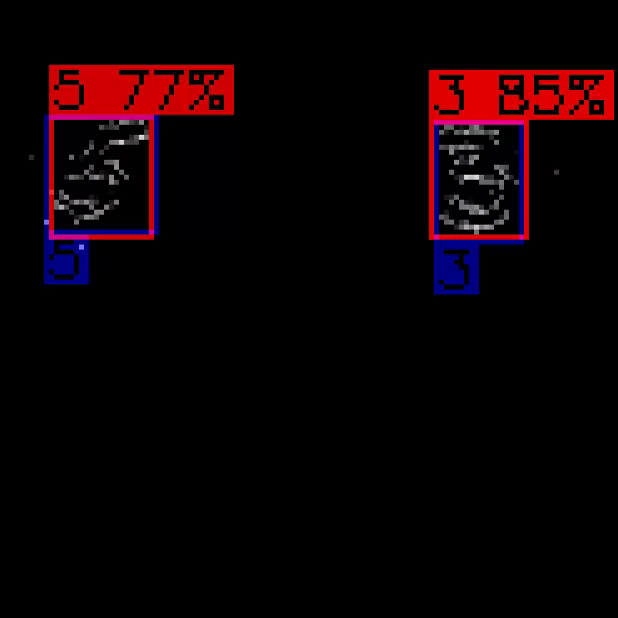}&
		\includegraphics[width=0.18\textwidth]{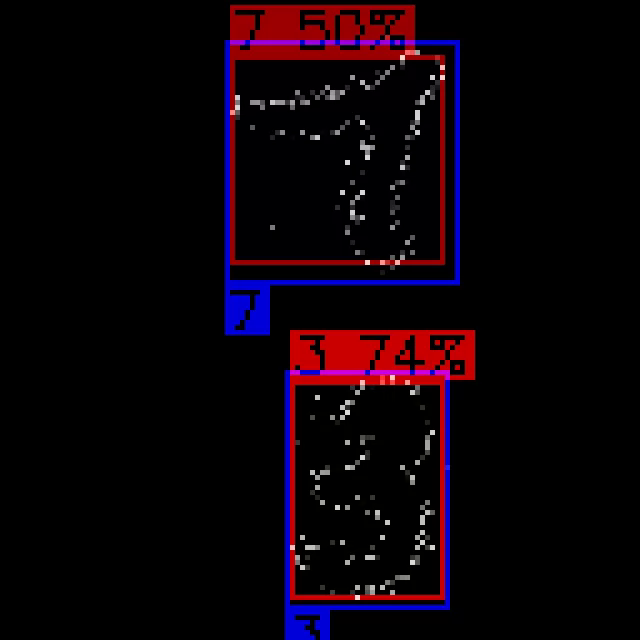}&
		\includegraphics[width=0.18\textwidth]{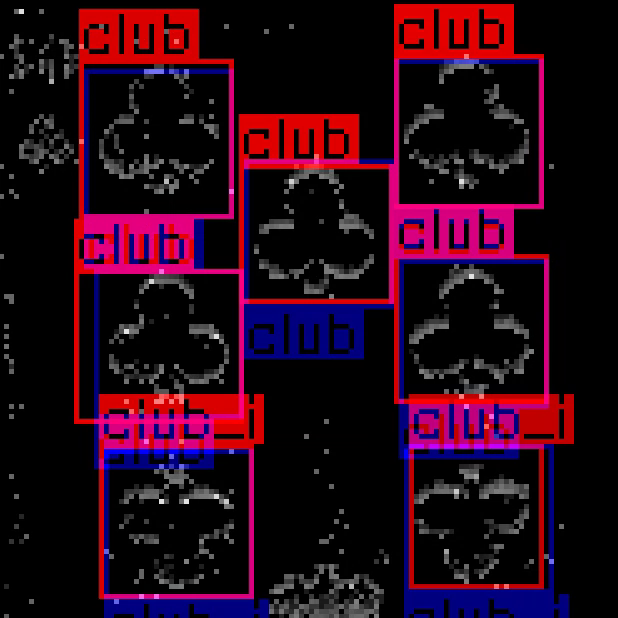}&
		\includegraphics[width=0.25\textwidth]{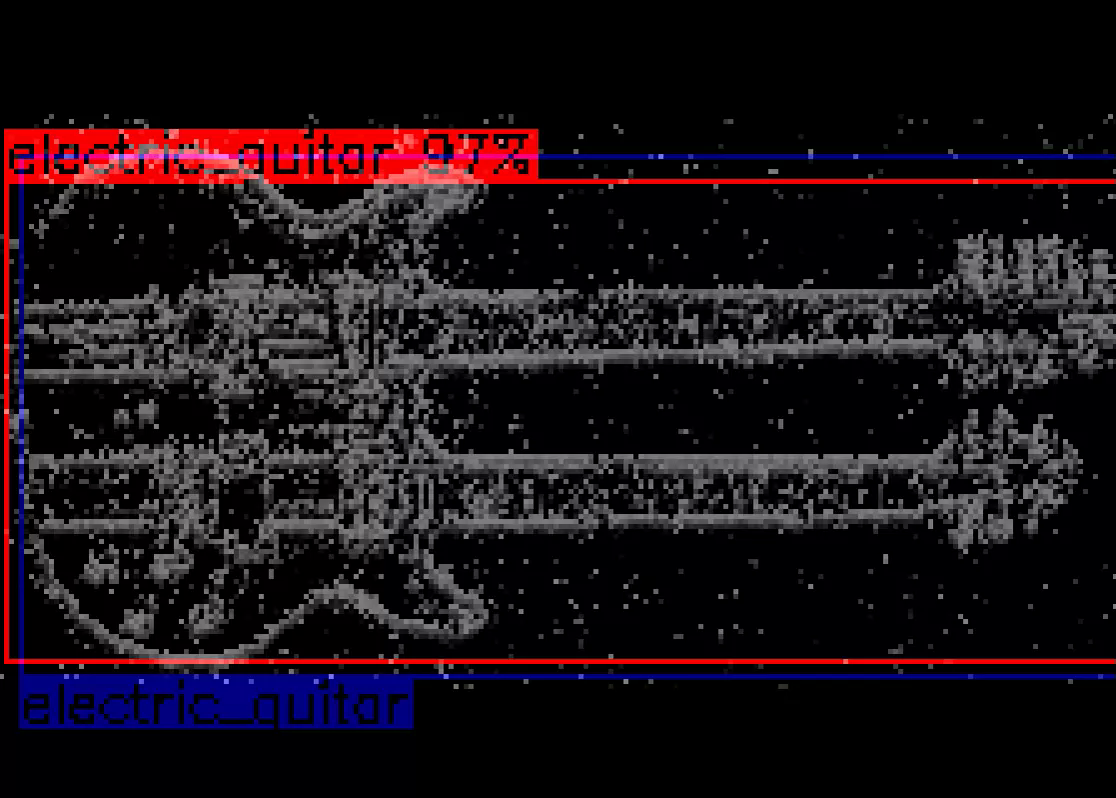}&
		\includegraphics[width=0.18\textwidth]{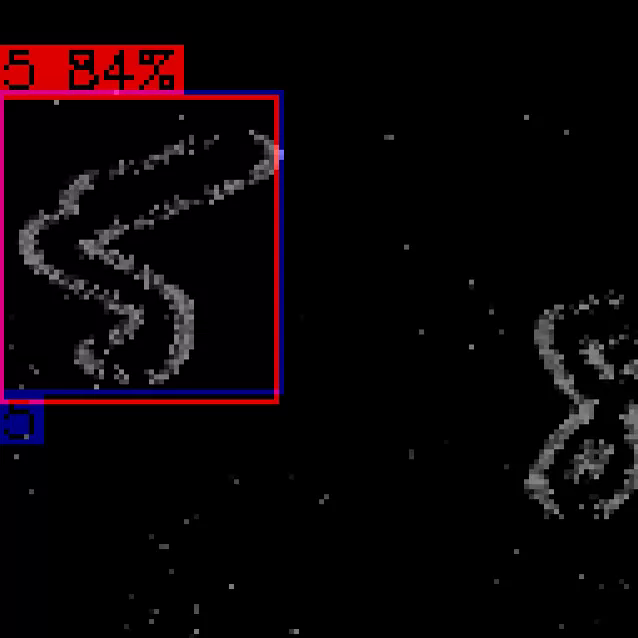}\\
		\includegraphics[width=0.18\textwidth]{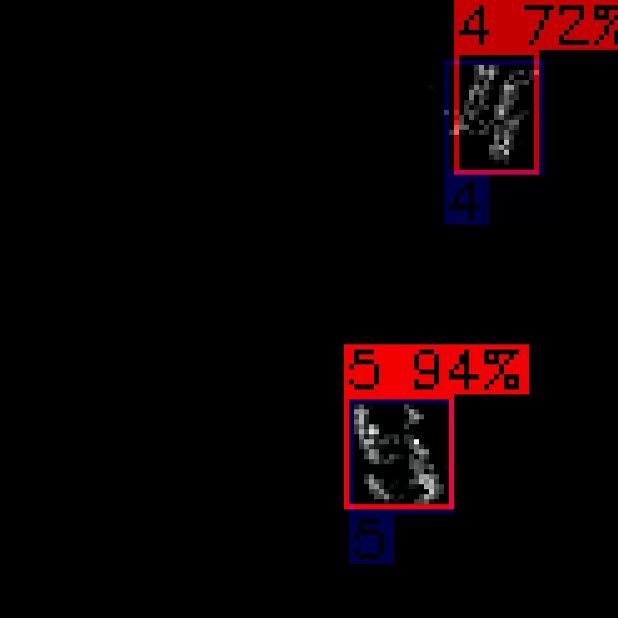}&
		\includegraphics[width=0.18\textwidth]{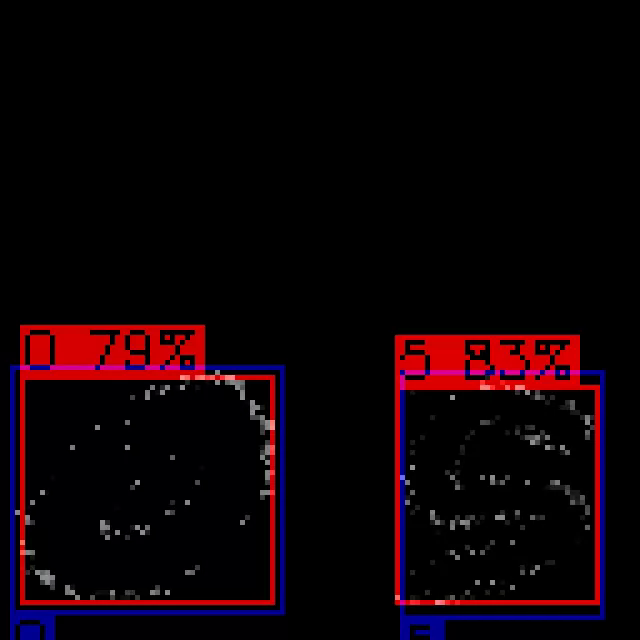}&
		\includegraphics[width=0.18\textwidth]{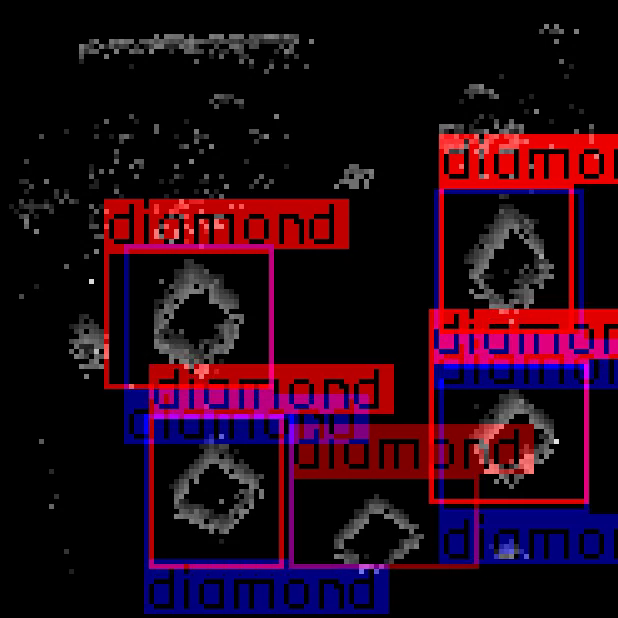}&
		\includegraphics[width=0.25\textwidth]{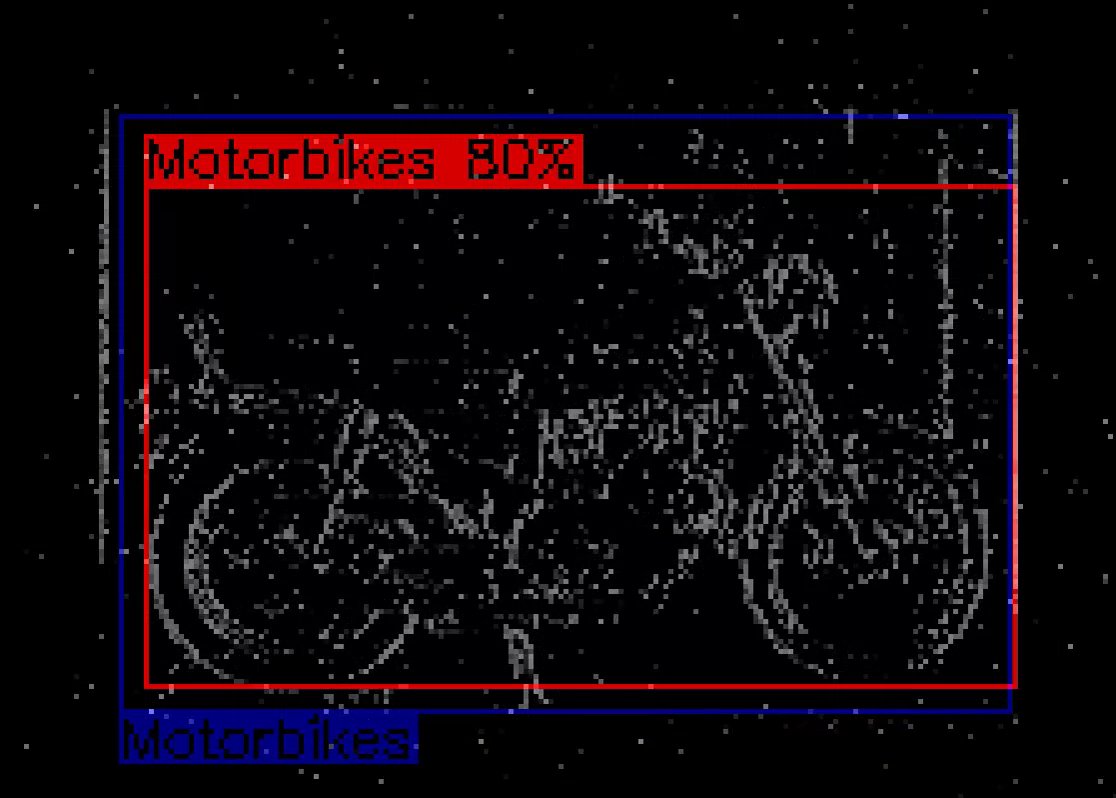}&
		\includegraphics[width=0.18\textwidth]{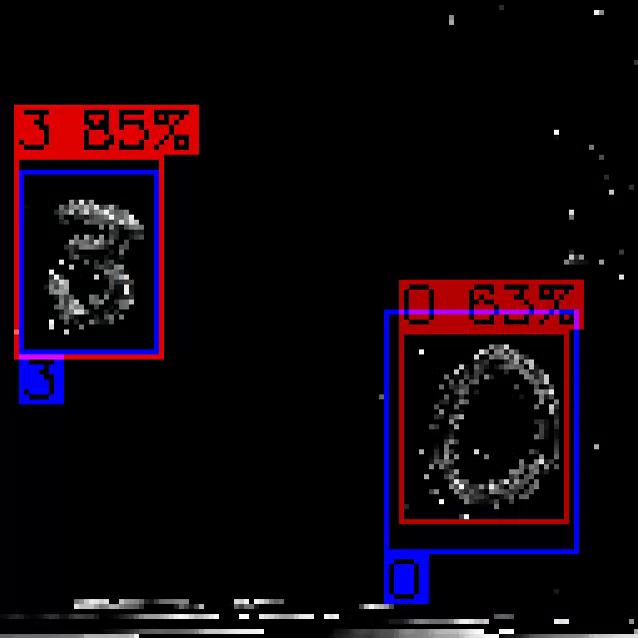}\\
	\end{tabular}
	\vspace{-8pt}
	\caption{Examples of YOLE predictions. }
	\vspace{-0.5cm}
	\label{fig:examples}
\end{figure*}

\paragraph{Detection performance of YOLE.}
The YOLE network achieves good detection results both in terms of mean average precision (mAP)~\citePaper{Everingham2010Jun} and accuracy, which in this case is computed by matching every ground truth bounding box with the predicted one having the highest intersection over union (IOU), in most of the datasets.
The results we obtained are summarized in Table~\ref{tab:results}.

We used the Shifted N-MNIST dataset also to analyze how detection performance changes when the network is used to process scenes composed of a variable number of objects, as reported in Table \ref{tab:resNMNIST}. We denote as \emph{v1} the results obtained using scenes composed of a single digit and with \emph{v2} those obtained with scenes containing two digits in random locations of the field of view. We further tested the robustness of the proposed model by adding some challenging noise. We added non-target objects (\emph{v2fr}) in the form of five $8 \times 8$ fragments, taken from random N-MNIST digits using a procedure similar to the one used to build the \emph{Cluttered Translated MNIST} dataset~\citePaper{Mnih2014Jun}, and $200$ additional random events per frame (\emph{v2fr+ns}).

In case of multiple objects the algorithm is still able to detect all of them, while, as expected, performance drops both in terms of accuracy and mean average precision when dealing with noisy data. 
Neverthelesss, we achieved very good detection performance on the Shifted MNIST-DVS, Blackboard MNIST and Poker-DVS datasets which represent a more realistic scenario in terms of noise.
All of these experiments were performed using the set of hyperparameters suggested by the original work from~\citePaper{Redmon2015Jun}. However, a different choice of these parameters, namely $\lambda_{coord} = 25.0$ and $\lambda_{noobj} = 0.25$, worked better for us increasing both the accuracy and mean average precision scores (\emph{v2*}).

The dataset in which the proposed model did not achieve noticeable results is N-Caltech101. This is mainly explained by the increased difficulty of the task and by the fact that the number of samples in each class is not evenly balanced. 
The network, indeed, usually achieves good results when the number of training samples is high such as with \emph{Airplanes}, \emph{Motorbikes} and \emph{Faces\_easy}, and in cases in which samples are very similar, \eg, \emph{inline\_skate} (see Table \ref{tab:YOLE_ncaltech101} and supplementary material). As the number of training samples decreases and the sample variability within the class increases, however, the performance of the model becomes worse, behavior which explains the poor aggregate scores we report in Table~\ref{tab:results}.

\paragraph*{Detection performance of fcYOLE.}
With this fully-convolutional variant of the network we registered a slight decrease in performance w.r.t. the results we obtained using YOLE, as reported in Table \ref{tab:results} and Table \ref{tab:fcYOLE_ncaltech101}. This gap in performance is mainly due to the fact that each region in fcYOLE generates its predictions by only looking at the visual information contained in its portion of the field of view. Indeed, if an object is only partially contained inside a region the network has to guess the object dimensions and class by only looking at a restricted region of the surface.
It should be stressed, however, that the difference in performance between the two architectures does not come from the use of the proposed event layers, whose output are the same as the conventional ones, but rather from the reduced expressive power caused by the absence of fully-connected layers in fcYOLE. Indeed, not removing them would have allowed us to obtain the same performance of YOLE, but with the drawback of being able to exploit event-based layers only up to the first FC-layer, which has not been formalized yet in an event-based form. Removing the last fully-connected layers allowed us to design a detection network made of only event-based layers and which uses also a significantly lower number of parameters.
In the supplementary materials we provide a video showing a comparison between YOLE and fcYOLE predictions.

To identify the advantages and weaknesses of the proposed event-based framework in terms of inference time we compared our detection networks on two datasets, Shifted N-MNIST and Blackboard MNIST. We group events into batches of $10$ms and average timings on $1000$ runs. In the first dataset the event-based approach achieved a $2$x speedup ($22.6$ms per batch), whereas in the second one it performed slightly slower ($43.2$ms per batch) w.r.t. a network making use of conventional layers ($34.6$ms per batch). The second benchmark is indeed challenging for our framework since changes are not localized in restricted regions.
Our current implementation is not optimized to handle noisy scenes efficiently. Indeed, additional experiments showed that asynchronous CNNs are able to provide a faster prediction only up to a $80\%$ of event sparsity (where with sparsity we mean the percentage of changed pixels in the reconstructed image). Further investigations are out of the scope of this paper and will be addressed in future works.


\section{Conclusions}
\label{sec:concl}
We proposed two different methods, based on the YOLO architecture, to accomplish object detection in event-based cameras. The first one, namely YOLE, integrates events into a unique leaky surface.
Conversely, fcYOLE relies on a more general extension of the convolutional and max pooling layers to directly deal with events and exploit their sparsity by preventing the whole network to be reprocessed. The obtained asynchronous detector dynamically adapts to the timing of the events stream by producing results only as a consequence of incoming events and by maintaining its internal state, without performing any additional computation, when no events arrive.
This novel event-based framework can be used in every fully-convolutional architecture to make it usable with event-cameras, even conventional CNN for classification, although in this paper it has been applied to object detection networks.

We analyzed the timing performance of this formalization obtaining promising results. We are planning to extend our framework to automatically detect at runtime when the use of event-based layers speeds up computation (\ie, changes affect few regions of the surface) or a complete recomputation of the feature maps is more beneficial in order to exploit the benefits of both approaches. Nevertheless, we believe that a ad-hoc hardware implementation, would allow to better exploit the advantages of the proposed method enabling a fair timing comparison with SNNs, which are usually implemented in hardware.

{\small
	\vspace{-8pt}
	\paragraph{Acknowledgements}
	We would like to thank Prophesee for helpful discussions on YOLE.
	The research leading to these results has received funding from project TEINVEIN: TEcnologie INnovative per i VEicoli Intelligenti, CUP (Codice Unico Progetto - Unique Project Code): E96D17000110009 - Call ``Accordi per la Ricerca e l’Innovazione", cofunded by POR FESR 2014-2020 (Programma Operativo Regionale, Fondo Europeo di Sviluppo Regionale – Regional Operational Programme, European Regional Development Fund).
	
\bibliographystylePaper{ieee}
\bibliographyPaper{egbib}
}

\clearpage
\pagebreak

\setcounter{section}{0}
\setcounter{figure}{0}
\setcounter{page}{1}
\title{Asynchronous Convolutional Networks for Object Detection \\ in Neuromorphic Cameras\\Supplementary material}

\author{Marco Cannici \hspace{20pt} 
	Marco Ciccone \hspace{20pt} 
	Andrea Romanoni\hspace{20pt} 
	Matteo Matteucci\\
	Politecnico di Milano, Italy\\
	{\tt\small \{marco.cannici,marco.ciccone,andrea.romanoni,matteo.matteucci\}@polimi.it}
}

\maketitle
\begin{figure*}[pt]
	\vspace{1cm}
	\begin{center}
		\setlength{\tabcolsep}{1px}
		\begin{tabular}{cccc}
			\hline
			\includegraphics[width=0.25\textwidth]{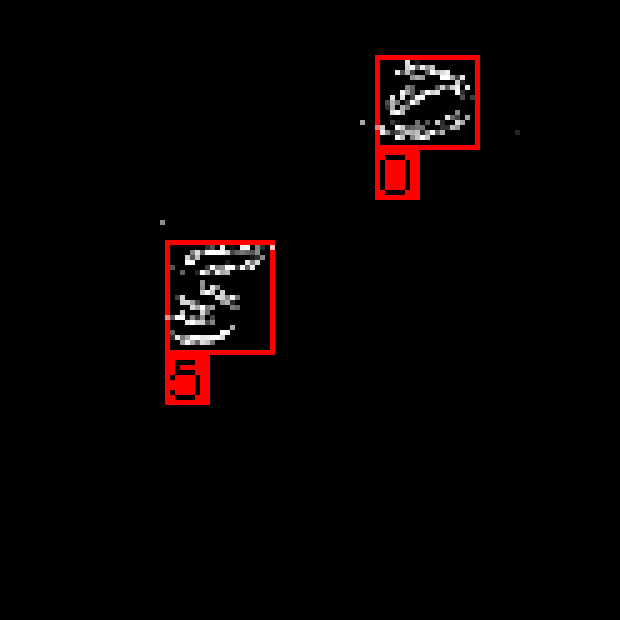}&
			\includegraphics[width=0.25\textwidth]{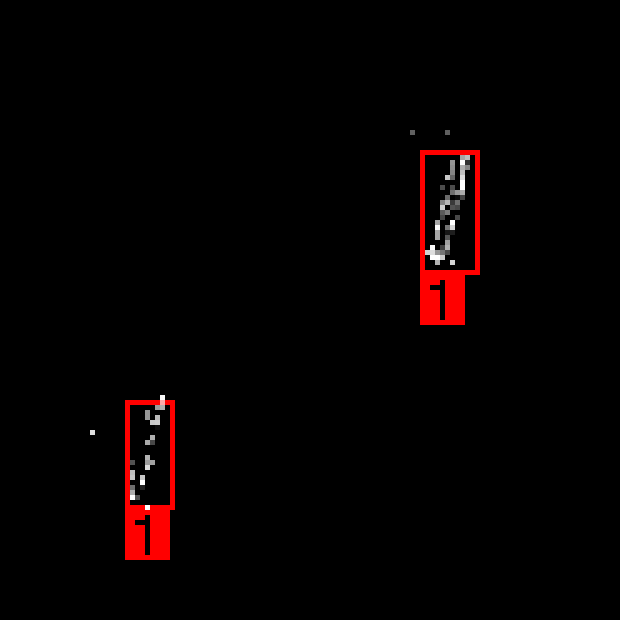}&
			\includegraphics[width=0.25\textwidth]{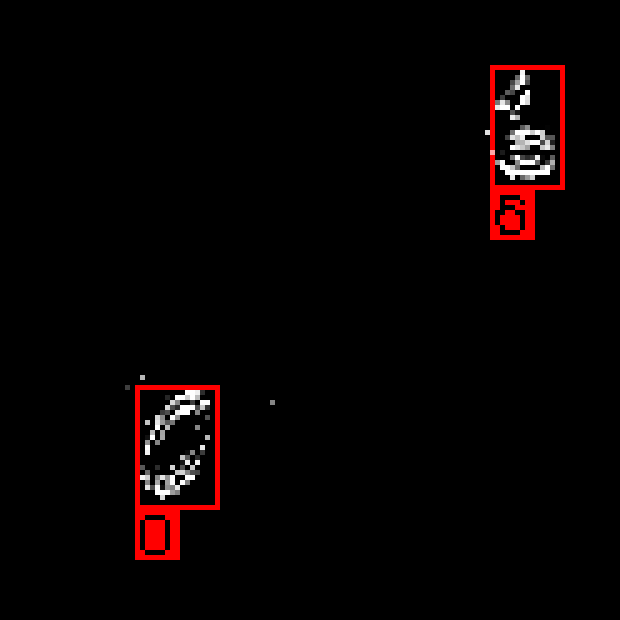}&
			\includegraphics[width=0.25\textwidth]{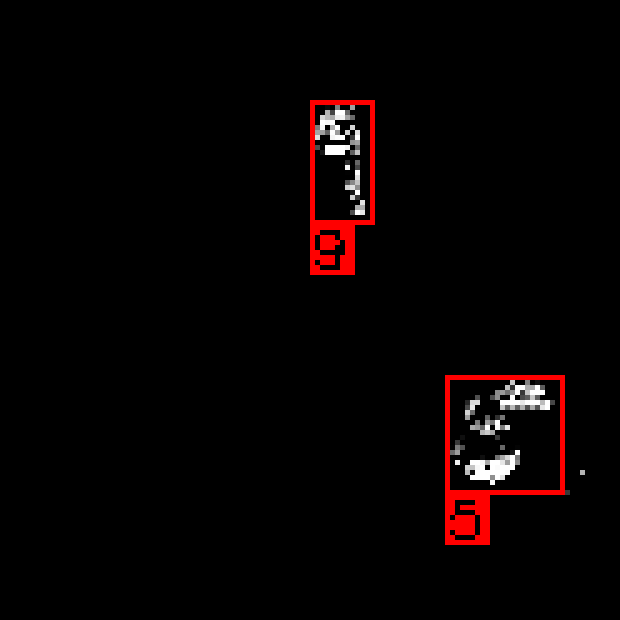}\\
			\multicolumn{4}{c}{Shifted N-MNIST}\\
			\hline
			\includegraphics[width=0.25\textwidth]{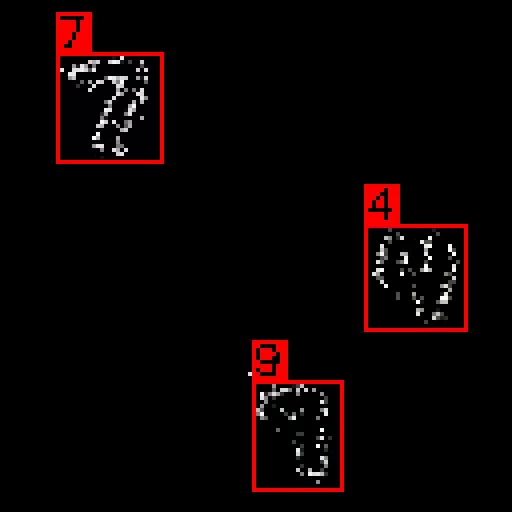}&
			\includegraphics[width=0.25\textwidth]{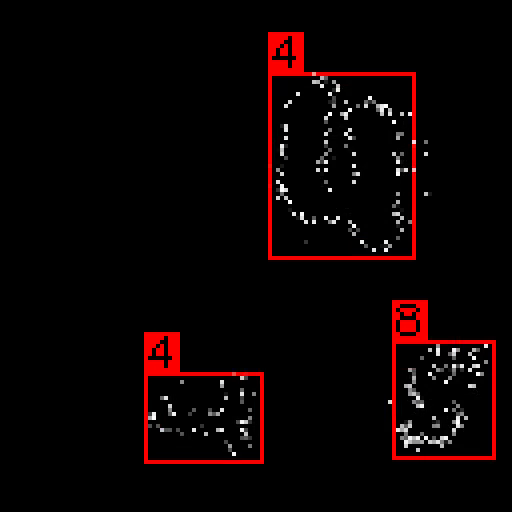}&
			\includegraphics[width=0.25\textwidth]{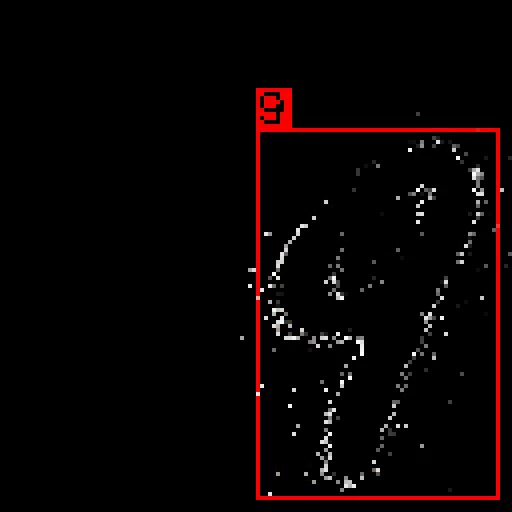}&
			\includegraphics[width=0.25\textwidth]{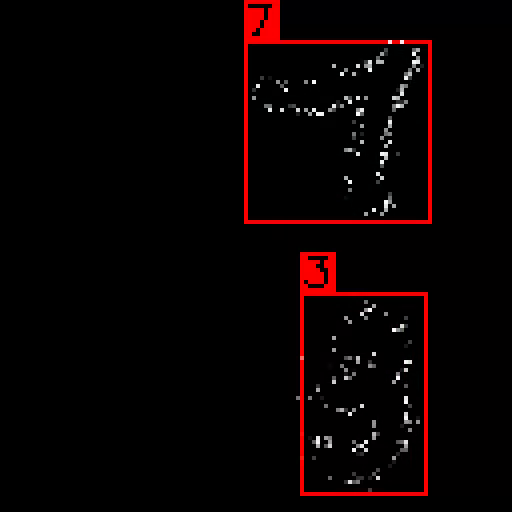}\\
			\multicolumn{4}{c}{Shifted MNIST-DVS}\\
			\hline
			\includegraphics[width=0.25\textwidth]{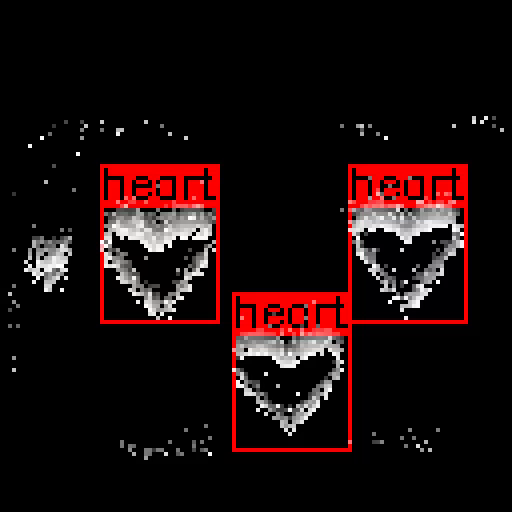}&
			\includegraphics[width=0.25\textwidth]{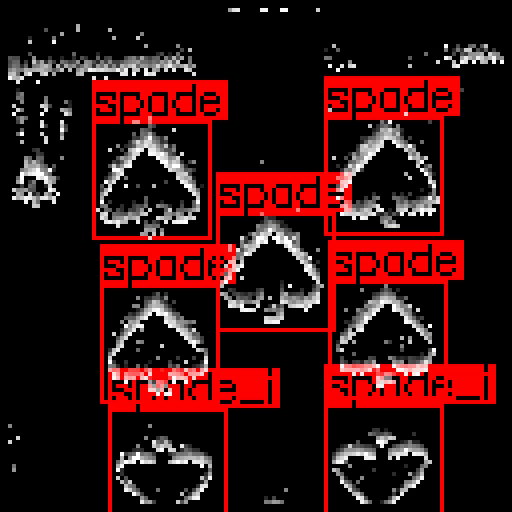}&
			\includegraphics[width=0.25\textwidth]{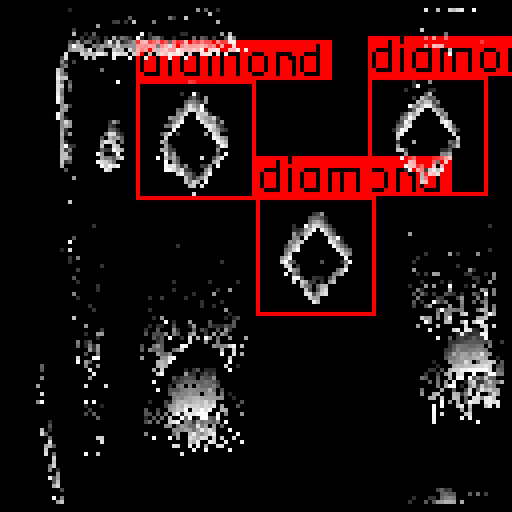}&
			\includegraphics[width=0.25\textwidth]{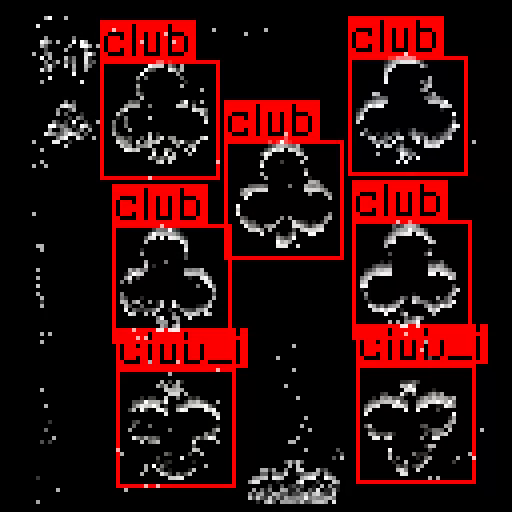}\\
			\multicolumn{4}{c}{OD-Poker-DVS}\\
			\hline
			\includegraphics[width=0.25\textwidth]{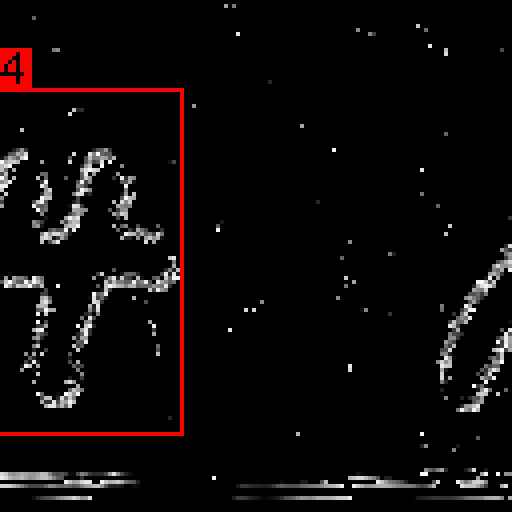}&
			\includegraphics[width=0.25\textwidth]{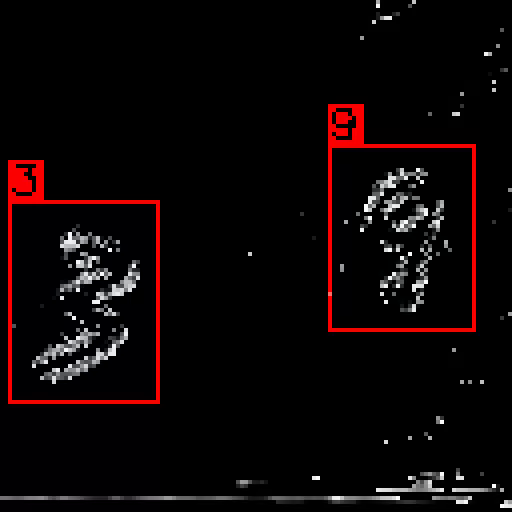}&
			\includegraphics[width=0.25\textwidth]{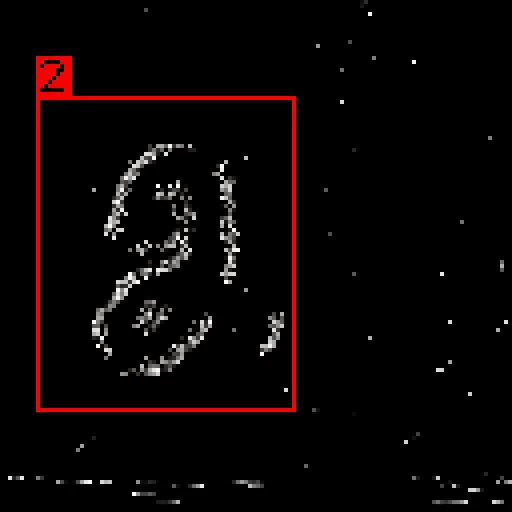}&
			\includegraphics[width=0.25\textwidth]{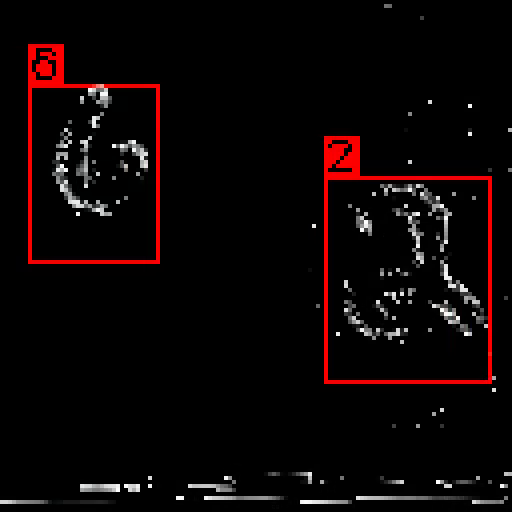}\\
			\multicolumn{4}{c}{Blackboard-MNIST}
		\end{tabular}
		\caption{Examples of samples from the proposed datasets. }
		\label{fig:examples}
	\end{center}
\end{figure*}

\begin{figure*}[pt]
	\begin{center}
		\setlength{\tabcolsep}{1px}
		\begin{tabular}{ccccc}
			\hline
			\includegraphics[width=0.25\textwidth]{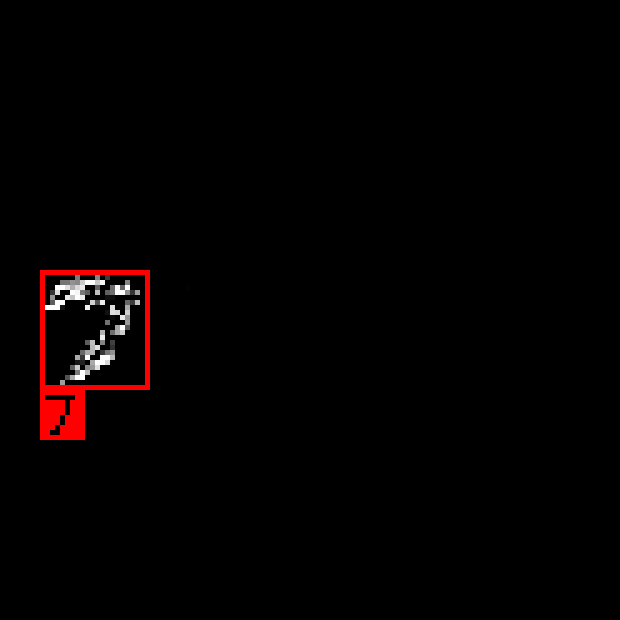}&
			\includegraphics[width=0.25\textwidth]{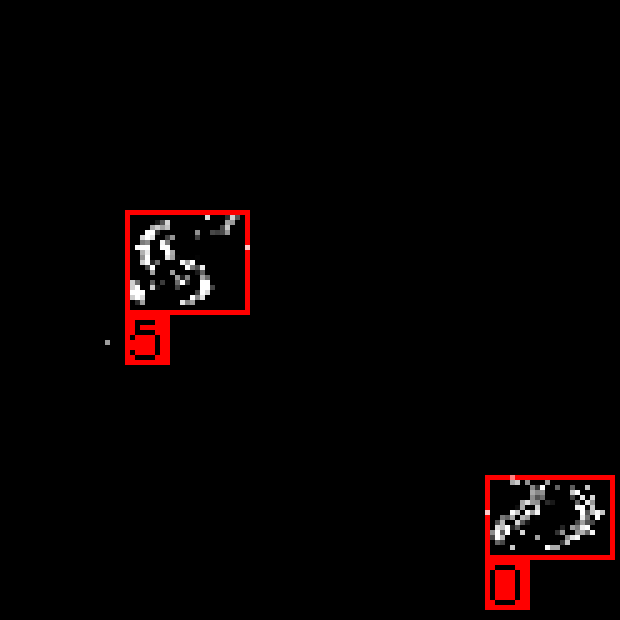}&
			\includegraphics[width=0.25\textwidth]{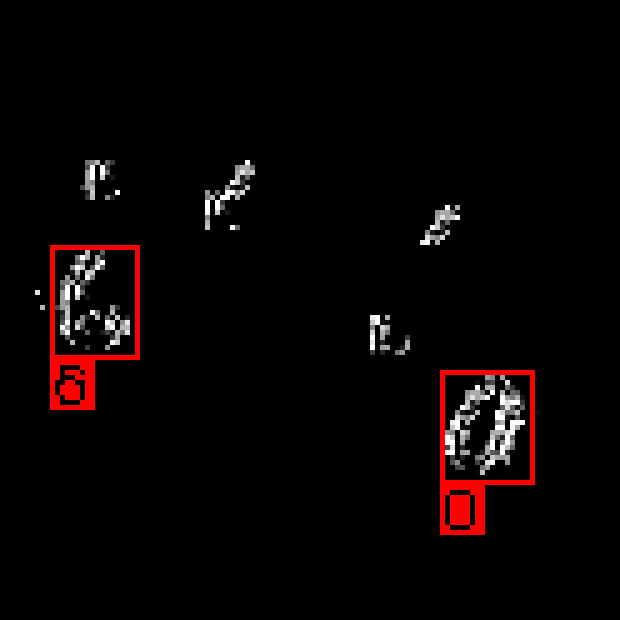}&
			\includegraphics[width=0.25\textwidth]{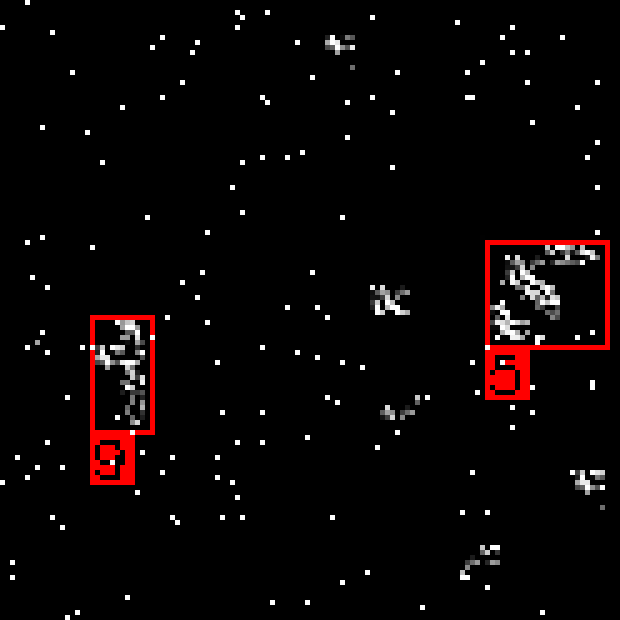}\\
			v1&v2&v2fr&v2fr+ns\\
		\end{tabular}
		\caption{Different versions of Shifted N-MNIST. }
		\label{fig:N-MNIST}
	\end{center}
\end{figure*}

\begin{figure*}[pt]
	\begin{center}
		\includegraphics[width=\textwidth]{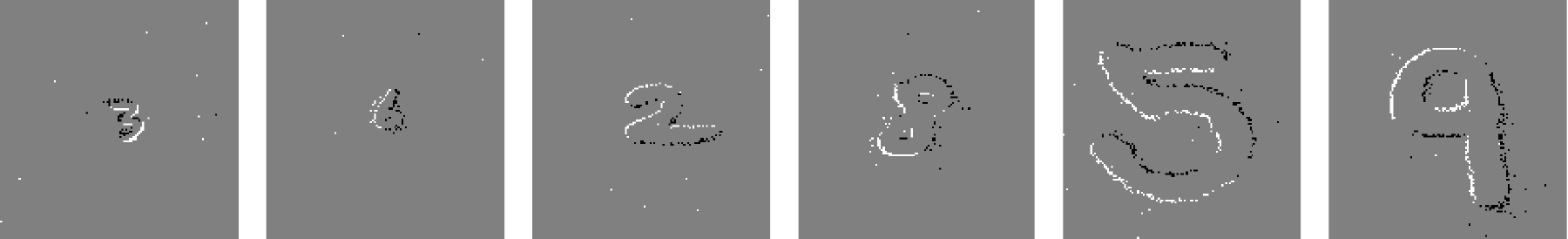}
		\caption{Examples of the three different scales of MNIST-DVS digits. Two samples at scale \emph{scale4}, two at \emph{scale8} and two at \emph{scale16}.}
		\label{fig:mnist-dvs_examples}
	\end{center}
\end{figure*}

In this document we describe our novel event-based datasets adopted in the paper ``Asynchronous Convolutional Network for Object Detection in Neuromorphic Cameras".

\section{Event-based object detection datasets} 
Due to the lack of object detection datasets with event cameras, we extended the publicly available N-MNIST, MNIST-DVS, Poker-DVS and we propose a novel dataset based on MNIST, i.e.,  Blackboard MNIST.
They will be soon released, however, in Figure \ref{fig:examples} we reported some example from the four datasets.
\subsection{Shifted N-MNIST} 
\label{sec:shifted_n-mnist}

The original N-MNIST \citeSuppl{Orchard2015Nov} extends the well-known MNIST \citeSuppl{Lecun1998Nov}: it provides an event-based representation of both the full training set ($60,000$ samples) and the full testing set ($10,000$ samples) to evaluate object classification algorithms.
The dataset has been recorded by means of event camera in front of an LCD screen and moved to detect static MNIST digits displayed on the monitor.
For further details we refer the reader to \citeSuppl{Orchard2015Nov}.

Starting from the N-MNIST dataset, we built a more complex set of recordings that we used 
to train the object detection network to detect multiple objects in the same scene.
We created two versions of the dataset, Shifted N-MNIST v1 and Shifted N-MNIST v2, that contains respectively one or two non overlapping $34 \times 34$ N-MNIST digits per sample randomly positioned on a bigger surface. 
We used different surface dimensions in our tests which vary from double the original size, $68 \times 68$, up to $124 \times 124$. The dimension and structure of the resulting dataset is the same of the original N-MNIST collection.

To extend the dataset for object detection evaluation, the bounding boxes ground truth are required.
To estimate them we first integrate events into a single frame as described in Section 2 of the original paper.
We remove the noise by considering only non-zero pixels having at least other $\rho$ non-zero pixels around them within a circle of radius $\mathit{R}$. All the other pixels are considered noise.
Then, with a custom version of the DBSCAN \citeSuppl{Ester1996} density-based clustering algorithm we group pixels into a single cluster. A threshold ${min}_{area}$ is used to filter out small bounding boxes extracted in correspondence of low events activities. This condition usually happens during the transition from a saccade to the next one as the camera remains still for a small fraction of time and no events are generated. We used $\rho = 3$, $\mathit{R} = 2$ and ${min}_{area} = 10$. The coordinates of these bounding boxes are then shifted based on the final position the digit has in the bigger field of view. 

For each N-MNIST sample, another digit was randomly selected in the same portion of the dataset (training, testing or validation) to form a new example. The final dataset contains $60,000$ training samples and $10,000$ testing samples, as for the original N-MNSIT dataset.
In Figure \ref{fig:N-MNIST} we illustrate one example for v1 and the three variants of v2 we adopted (and described) in the paper.

\subsection{Shifted MNIST-DVS} 
\label{sec:shifted_mnist-dvs}

The MNIST-DVS dataset \citeSuppl{Serrano-Gotarredona2015Dec} is another collection of event-based recordings that extends  MNIST \citeSuppl{Lecun1998Nov}.
It consists of $30,000$ samples recorded by displaying digits on an  screen in front of a event camera, but differently from N-MNIST, they move digits  on the screen instead of the sensors, and they use the digits at three different scales, i.e., \textit{scale4}, \textit{scale8} and \textit{scale16}.
The resulting dataset is composed of $30,000$ event-based recordings showing each one of the selected $10,000$ MNIST digits on thee different dimensions. Examples of these recordings are shown in Figure \ref{fig:mnist-dvs_examples}.

We used MNIST-DVS recordings to build a detection dataset by means of a procedure similar to the one we used to create the Shifted N-MNIST dataset. 
However in this case we mix together digits of multiple scales.
All the MNIST-DVS samples, despite of the actual dimensions of the digits being recorded, are contained within a fixed $128 \times 128$ field of view. Digits are placed centered inside the scene and occupy a limited portion of the frame, especially those belonging to the smallest and middle scales. In order to place multiple examples on the same scene we first cropped the three scales of samples into smaller recordings occupying $35 \times 35$, $65 \times 65$ and $105 \times 105$ spatial regions respectively. The bounding boxes annotations and the final examples were obtained by means of the same procedure we used to construct the Shifted N-MNIST dataset. 
These recordings were built by mixing digits of different dimensions in the same sample. Based on the original samples dimensions, we decided to use the following four configurations (which specify the number of samples of each category used to build a single Shifted MNIST-DVS example):  (i) three \emph{scale4} digits, (ii) two \emph{scale8} digits, (iii) two \emph{scale4} digits mixed with one \emph{scale8} digit (iv) one \emph{scale16} digit placed in random locations of the field of view.
The overall dataset is composed of $30,000$ samples containing these four possible configurations.

\subsection{OD-Poker-DVS} 
\label{sec:poker-dvs}
The original Poker-DVS \citeSuppl{Serrano-Gotarredona2015Dec} have been proposed to test object recognition algorithms; it is a small collection of neuromorphic recordings obtained by quickly browsing custom made poker card decks in front of a DVS camera for 2-4 seconds.
The dataset is composed of $131$ samples containing centered pips of the four possible categories (spades, hearts, diamonds or clubs) extracted from three decks recordings. Single pips were extracted by means of an event-based tracking algorithms which was used to follow symbols inside the scene and to extract $31 \times 31$ pixels examples. 

With OD-Poker-DVS we extend its scope to test also object detection.
To do so we used the event-based tracking algorithm provided with the original dataset to follow the movement of the $31 \times 31$ samples in the uncut recordings and extract their bounding boxes. The final dataset was obtained using a procedure similar to the one used in \citeSuppl{Stromatias2017Jun}. Indeed, we divided the sections of the three original decks recordings containing visible digits into a set of shorter examples, each of which about $1.5$ms long. Examples were split in order to ensure approximately the same number of objects (i.e., ground truth bounding boxes) in each example. The final detection dataset is composed of $292$ small examples which we divided into $218$ training and $74$ testing samples.

Even if composed of a limited amount of samples, this dataset represents an interesting real-world application that highlights the potential of event-based vision sensors. The nature of the data acquisition is indeed particularly well suited to neuromorphic cameras due to their very high temporal resolution. Symbols are clearly visible inside the recordings even if they move at very high speed. Each pip, indeed, takes from $10$ to $30$ ms to cross the screen but it can be easily recognized within the first $1$-$2$ ms.

\subsection{Blackboard MNIST} 
\label{sec:mnist-sim}

The two dataset based on MNIST presented in Section \ref{sec:shifted_n-mnist} and \ref{sec:shifted_mnist-dvs} have the drawback of recording digits at predefined sizes. 
Therefore, in Blackboard MNIST we propose a more challenging scenario that consists of a number of samples showing digits (from the original MNIST dataset) written on a blackboard in random positions and with different scales.

\begin{figure*}[tp]
	\centering
	\begin{subfigure}[c]{0.6\textwidth}
		\begin{center}
			\centering
			\includegraphics[height=0.5\textwidth]{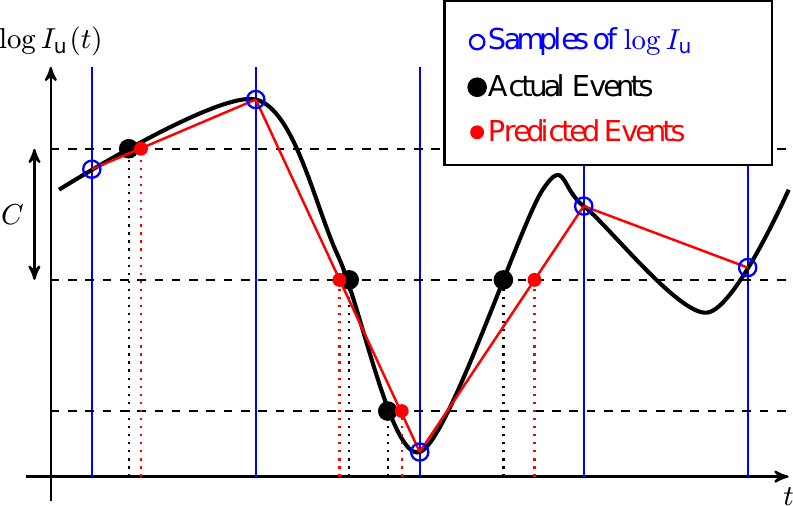}
			\caption{} \label{fig:davis_sim_pixel}    
		\end{center}
	\end{subfigure}%
	\hfill
	\begin{subfigure}[c]{0.4\textwidth}
		\centering
		\includegraphics[height=0.7\textwidth]{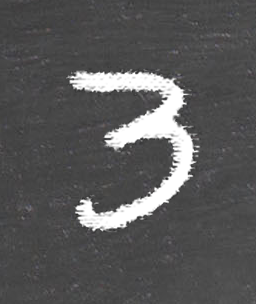}
		\caption{} \label{fig:blackboard-mnist_digit}
	\end{subfigure}%
	\caption[]{\emph{(a)} The image shows in black the intensity, expresses as $\log I_{{u}}(t)$, of a single pixel $\mathbf{u}=(x,y)$. This curve is sampled at a constant rate when frames are generated by Blender, shown in figure as vertical blue lines. The sampled values thus obtained (blue circles) are used to approximate the pixel intensity by means of a simple piecewise linear time interpolation (red line). Whenever this curve crosses one of the threshold values (horizontal dashed lines) a new event is generated with the corresponding predicted timestamp. (Figure from \citeSuppl{Mueggler2016Oct}) \textbf{(b)} A preprocessed MNIST digit on top of the blackboard's background.}
\end{figure*}

We used the DAVIS simulator released by \citeSuppl{Mueggler2016Oct} to build our own set of synthetic recordings. Given a three-dimensional virtual scene and the trajectory of a moving camera within it,  the simulator is able to generate a stream of events describing the visual information captured by the virtual camera. The system uses Blender \citeSuppl{Blender2017}, an open-source 3D modeling tool, to generate thousands of rendered frames along a predefined camera trajectory which are then used to reconstruct the corresponding neuromorphic recording. The intensity value of each single pixel inside the sequence of rendered frames, captured at a constant frame-rate, is tracked. As Figure \ref{fig:davis_sim_pixel} shows, an event is generated whenever the log-intensity of a pixel crosses an intensity threshold, as in a real event-based camera. A piecewise linear time interpolation mechanism is used to determine brightness changes in the time between frames in order to simulate the microseconds timestamp resolution of a real sensor.
We extended the simulator to output bounding boxes annotations associated to every visible digit.

We used Blender APIs to place MNIST digits in random locations of a blackboard and to record their position with respect to the camera point of view.
Original MNIST images depict black handwritten digits on a white background. 
To mimic the chalk on the blackboard, we removed the background, we turned digits in white and we roughen their contours to make them look like if their were written with a chalk. 
An example is shown in Figure \ref{fig:blackboard-mnist_digit}.

\begin{figure*}[tp]
	\centering
	\begin{subfigure}[c]{0.58\textwidth}
		\begin{center}
			\includegraphics[height=0.45\textwidth]{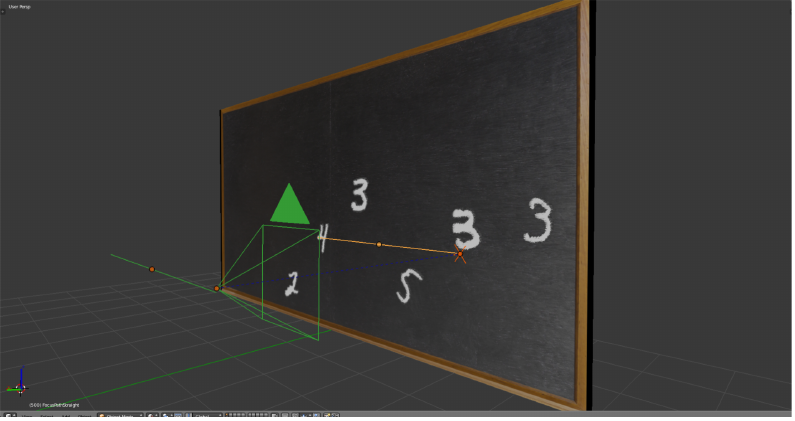}
			\caption{} \label{fig:blackboard-dvs_scene}    
		\end{center}
	\end{subfigure}%
	\hfill
	\begin{subfigure}[c]{0.42\textwidth}
		\begin{center}
			\includegraphics[height=0.55\textwidth]{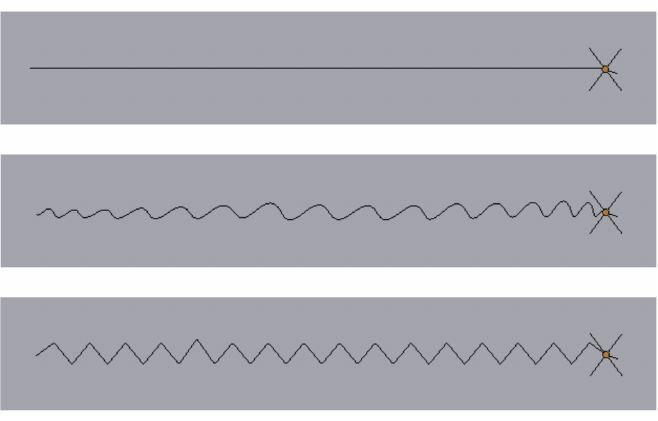}
			\caption{} \label{fig:blackboard-dvs_scene_traj}
		\end{center}
	\end{subfigure}%
	\caption{\textbf{(a)} The 3D scene used to generate the Blackboard MNIST dataset. The camera moves in front of the blackboard along a straight trajectory while following a \emph{focus object} that moves on the blackboard's surface, synchronized with the camera. The camera and its trajectory are depicted in green, the focus object is represented as a red cross and, finally, its trajectory is depicted as a yellow line. \textbf{(b)} The three types of focus trajectories.}
	\label{fig:blackboard-dvs_scene}
\end{figure*}

The scene contains the image of a blackboard  on a vertical plane and a virtual camera with $128 \times 128$ resolution that moves horizontally on a predefined trajectory parallel to the blackboard plane (see Figure \ref{fig:blackboard-dvs_scene}). The camera points a hidden object that moves on the blackboard surface, synchronized with the camera, following a given trajectory. To introduce variability in the camera movement, and to allow all the digits outline to be seen (and possibly detected), we used different trajectories that vary from a straight path to a smooth or triangular undulating path that makes the camera tilt along the transverse axis while moving (Figure \ref{fig:blackboard-dvs_scene_traj}). 

Before starting the simulation, we randomly select a number of preprocessed MNIST digits and place them in a random portion of the blackboard. 
The camera moves so that all the digits will be framed during the camera movement. The simulation is then finally started on this modified scene to generate neuromorphic recordings. Every time a frame is rendered during the simulation, the bounding boxes of all the visible digits inside the frame are also extracted. This operation is performed by computing the camera space coordinates (or normalized device coordinates) of the top-left and bottom-right vertex of all the images inside the field of view. Since images are slightly larger than the actual digits they contain, we cropped every bounding box  to better enclose each digit and also to compensate the small offset in the pixels position introduced by the camera motion and by the linear interpolation mechanism. In addition, bounding boxes corresponding to objects which are only partially visible are also filtered out. In order to build the final detection dataset, this generation process is executed multiple times, each time with different digits.

We built three sub-collections of recordings with increasing level of complexity which we merged together to obtain our final dataset: \emph{Blackboard MNIST EASY}, \emph{Blackboard MNIST MEDIUM}, \emph{Blackboard MNIST HARD}. 
In Blackboard MNIST EASY, we used digits of only one dimension (roughly corresponding to the middle scale of MNIST-DVS samples) and a single type of camera trajectory which moves the camera from right to left with the focus object moving in a straight line. In addition, only three objects were placed on the blackboard using only a fixed portion of its surface. We collected a total of $1,200$ samples ($1,000$ training, $100$ testing, $100$ validation).  

Blackboard MNIST MEDIUM features more variability in the number and dimensions of the digits and in the types of camera movements. Moreover, the portion of the blackboard on which digits were added varies and may cover any region of the blackboard, even those near its edges. The camera motions were also extended to the set of all possible trajectories that combine either left-to-right or right-to-left movements with variable paths of the focus object. We used three types of trajectories for this object: a straight line, a triangular path or a smooth curved trajectory, all parallel to the camera trajectory and placed around the position of the digits on the blackboard. One of these path was selected randomly for every generated sample. Triangular and curved trajectories were introduced as we noticed that sudden movements of the camera produce burst of events that we wanted our detection system to be able to handle. The number and dimensions of the digits were chosen following three possible configurations, similarly to the Shift MNIST-DVS dataset: either six small digits (with sizes comparable to \emph{scale4} MNIST-DVS digits), three intermediate-size digits (comparable to the MNIST-DVS \emph{scale8}) or two big digits (comparable to the biggest scale of the MNIST-DVS dataset, \emph{scale16}). A set of $1,200$ recordings was generated using the same splits of the first variant and with equal amount of samples in each one of the three configurations.

Finally, Blackboard MNIST HARD  contains digits recorded by using the second and third configuration of objects we described previously. 
However, in this case each image was  resized to a variable size spanning from the original configuration size down to the previous scale. A total of $600$ new samples ($500$ training, $50$ testing, $50$ validation) were generated, $300$ of them containing three digits each and the remaining $300$ consisting of two digits with variable size.

\begin{figure*}[tp]
	\centering
	\includegraphics[width=0.95\textwidth]{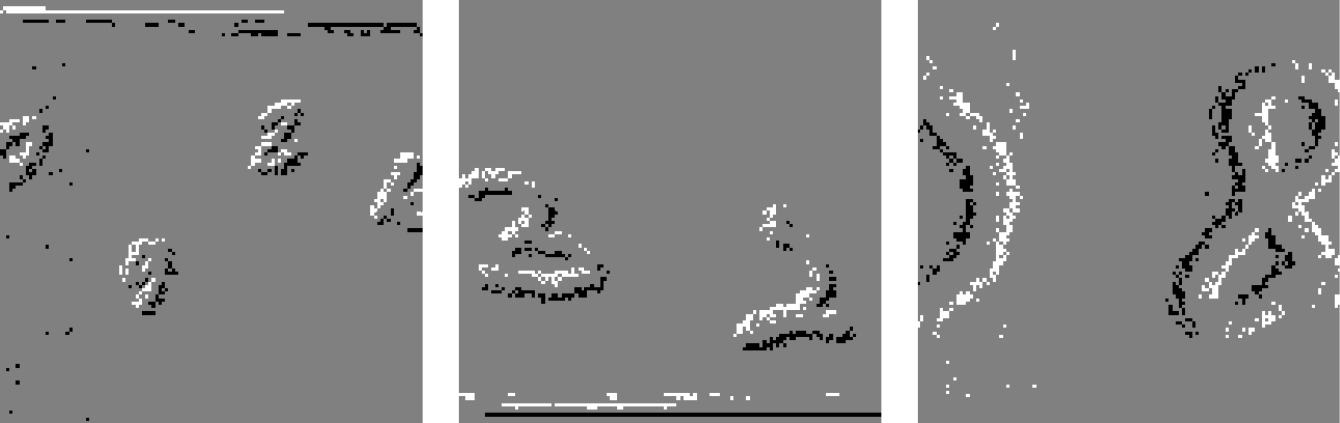}
	\caption{Examples of the three types of objects configurations used to generate the second collection of the Blackboard MNIST dataset.}
	\label{fig:blackboard-mnist-examples}
\end{figure*}

The three collections can be used individually or jointly; the whole Blackboard MNIST dataset contains $3,000$ samples in total ($2500$ training, $250$ testing, $250$ validation). Examples of different objects configurations are shown in Figure \ref{fig:blackboard-mnist-examples}. Samples were saved by means of the AEDAT v$3.1$ file format for event-based recordings.

\section{Results}

Table \ref{tab:ncaltech101_res} provides a comparison between the average precision of YOLE and fcYOLE on N-Caltech101 classes. We also provide a qualitative comparison between the two models in the video attachment.

\begin{table*}[!htbp]
	\caption{YOLE and fcYOLE average precisions on N-Caltech101}
	\label{tab:ncaltech101_res}
	\centering
	\scriptsize
	\begin{tabularx}{\textwidth}{c|*{20}{X}}
		& \rot{Motorbikes} & \rot{airplanes} & \rot{Faces\_easy} & \rot{watch} & \rot{Leopards} & \rot{chair} & \rot{bonsai} & \rot{car\_side} & \rot{ketch} & \rot{flamingo} & \rot{ant} & \rot{chandelier} & \rot{crocodile} & \rot{grand\_piano} & \rot{brain} & \rot{hawksbill} & \rot{butterfly} & \rot{helicopter} & \rot{menorah} & \rot{starfish} \\
		\midrule 
		AP fcYOLE & 97.5 & 96.8 & 92.2 & 75.7 & 57.2 & 7.5 & 30.2 & 70.3 & 42.3 & 2.3 & 2.4 & 34.8 & 0.0 & 69.5 & 35.3 & 19.6 & 33.5 & 8.6 & 67.7 & 23.2 \\
		
		AP YOLE & 97.8 & 95.8 & 94.7 & 84.2 & 62.9 & 17.3 & 59.3 & 61.7 & 52.9 & 10.0 & 25.8 & 55.7 & 1.6 & 81.3 & 53.3 & 29.1 & 46.3 & 14.9 & 80.7 & 32.7 \\
		
		$N_{train}$ & 480 & 480 & 261 & 145 & 120 & 109 & 78 & 75 & 70 & 68 & 66 & 65 & 61 & 61 & 60 & 60 & 55 & 54 & 53 & 52 \\
		
		& \rot{scorpion} & \rot{kangaroo} & \rot{trilobite} & \rot{sunflower} & \rot{buddha} & \rot{ewer} & \rot{revolver} & \rot{laptop} & \rot{llama} & \rot{ibis} & \rot{minaret} & \rot{umbrella} & \rot{crab} & \rot{electric\_guitar} & \rot{cougar\_face} & \rot{dragonfly} & \rot{crayfish} & \rot{dalmatian} & \rot{ferry} & \rot{euphonium} \\
		\midrule 
		
		AP fcYOLE & 2.5 & 3.4 & 41.2 & 29.1 & 46.5 & 35.3 & 20.3 & 40.0 & 1.4 & 1.5 & 59.5 & 61.0 & 5.0 & 23.2 & 21.8 & 55.6 & 7.3 & 24.9 & 29.7 & 43.5 \\
		
		AP YOLE & 6.9 & 5.0 & 62.5 & 43.3 & 57.2 & 51.3 & 57.4 & 88.1 & 10.2 & 6.5 & 81.3 & 85.9 & 19.5 & 29.7 & 39.8 & 59.9 & 9.5 & 33.0 & 34.0 & 53.6 \\
		
		$N_{train}$ & 52 & 52 & 52 & 51 & 51 & 51 & 50 & 49 & 48 & 48 & 46 & 45 & 45 & 45 & 43 & 42 & 42 & 41 & 41 & 40 \\
		
		& \rot{lotus} & \rot{stop\_sign} & \rot{joshua\_tree} & \rot{soccer\_ball} & \rot{elephant} & \rot{schooner} & \rot{dolphin} & \rot{lamp} & \rot{stegosaurus} & \rot{rhino} & \rot{wheelchair} & \rot{cellphone} & \rot{yin\_yang} & \rot{cup} & \rot{sea\_horse} & \rot{pyramid} & \rot{windsor\_chair} & \rot{hedgehog} & \rot{bass} & \rot{nautilus} \\
		\midrule 
		
		AP fcYOLE & 18.1 & 55.1 & 30.2 & 57.3 & 11.4 & 37.3 & 6.5 & 17.7 & 34.5 & 5.8 & 25.8 & 46.5 & 60.4 & 5.6 & 1.9 & 41.1 & 52.3 & 13.3 & 4.2 & 13.0 \\
		
		AP YOLE & 27.6 & 61.7 & 29.8 & 51.5 & 6.0 & 56.8 & 11.5 & 45.3 & 44.6 & 11.3 & 25.3 & 54.6 & 63.3 & 17.5 & 8.8 & 48.6 & 65.2 & 9.8 & 4.0 & 50.4 \\
		
		$N_{train}$ & 40 & 40 & 40 & 40 & 40 & 39 & 39 & 37 & 37 & 37 & 37 & 37 & 36 & 35 & 35 & 35 & 34 & 34 & 34 & 33 \\
		
		& \rot{pizza} & \rot{emu} & \rot{accordion} & \rot{dollar\_bill} & \rot{tick} & \rot{crocodile\_head} & \rot{gramophone} & \rot{rooster} & \rot{camera} & \rot{pagoda} & \rot{cougar\_body} & \rot{barrel} & \rot{ceiling\_fan} & \rot{beaver} & \rot{cannon} & \rot{mandolin} & \rot{flamingo\_head} & \rot{brontosaurus} & \rot{stapler} & \rot{pigeon} \\
		\midrule 
		
		AP fcYOLE & 17.4 & 5.6 & 48.3 & 74.4 & 28.2 & 9.8 & 30.6 & 26.6 & 15.1 & 34.0 & 0.0 & 30.7 & 40.8 & 0.5 & 0.0 & 6.6 & 2.7 & 0.2 & 46.0 & 6.2 \\
		
		AP YOLE & 54.5 & 5.0 & 52.7 & 86.5 & 55.2 & 10.0 & 48.4 & 64.5 & 32.7 & 33.3 & 12.2 & 41.2 & 50.1 & 0.0 & 0.3 & 43.4 & 9.3 & 25.8 & 68.1 & 43.1 \\
		
		$N_{train}$ & 33 & 33 & 33 & 32 & 31 & 31 & 31 & 31 & 30 & 29 & 29 & 29 & 29 & 28 & 27 & 27 & 27 & 27 & 27 & 27 \\
		
		& \rot{headphone} & \rot{anchor} & \rot{scissors} & \rot{wrench} & \rot{okapi} & \rot{lobster} & \rot{panda} & \rot{saxophone} & \rot{mayfly} & \rot{water\_lilly} & \rot{garfield} & \rot{wild\_cat} & \rot{gerenuk} & \rot{platypus} & \rot{binocular} & \rot{octopus} & \rot{strawberry} & \rot{snoopy} & \rot{metronome} & \rot{inline\_skate} \\
		\midrule 
		
		AP fcYOLE & 10.7 & 14.6 & 28.2 & 14.7 & 10.0 & 0.0 & 4.7 & 59.7 & 0.5 & 3.1 & 23.4 & 0.0 & 4.8 & 13.1 & 0.4 & 14.0 & 0.5 & 8.3 & 63.4 & 37.2 \\
		
		AP YOLE & 21.1 & 17.3 & 47.5 & 29.7 & 44.6 & 0.0 & 12.2 & 68.4 & 0.7 & 14.7 & 62.3 & 0.0 & 7.2 & 34.7 & 11.8 & 13.8 & 29.4 & 53.1 & 88.3 & 75.1 \\
		
		$N_{train}$ & 26 & 26 & 25 & 25 & 25 & 25 & 24 & 24 & 24 & 23 & 22 & 22 & 22 & 22 & 21 & 21 & 21 & 21 & 20 & 19 \\
	\end{tabularx}
\end{table*}

\newpage

{\small
\bibliographystyleSuppl{ieee}
\bibliographySuppl{egbib}
}


\begin{thebibliography}{10}\itemsep=-1pt

\bibitem{AlexZihaoZhu_2018}
L.~Y. Alex Zihao~Zhu.
\newblock Ev-flownet: Self-supervised optical flow estimation for event-based
  cameras.
\newblock {\em Robotics: Science and Systems}, Jan 2018.

\bibitem{bardow2016simultaneous}
P.~Bardow, A.~J. Davison, and S.~Leutenegger.
\newblock Simultaneous optical flow and intensity estimation from an event
  camera.
\newblock In {\em Proceedings of the IEEE Conference on Computer Vision and
  Pattern Recognition}, pages 884--892, 2016.

\bibitem{Berner2013}
R.~Berner, C.~Brandli, M.~Yang, S.-C. Liu, and T.~Delbruck.
\newblock A $240 \times 180$ $10$mw $12$us latency sparse-output vision sensor
  for mobile applications.
\newblock pages C186--C187, 01 2013.

\bibitem{Cannici2019}
M.~{Cannici}, M.~{Ciccone}, A.~{Romanoni}, and M.~{Matteucci}.
\newblock Attention mechanisms for object recognition with event-based cameras.
\newblock In {\em 2019 IEEE Winter Conference on Applications of Computer
  Vision (WACV)}, pages 1127--1136, Jan 2019.

\bibitem{cao2015spiking}
Y.~Cao, Y.~Chen, and D.~Khosla.
\newblock Spiking deep convolutional neural networks for energy-efficient
  object recognition.
\newblock {\em International Journal of Computer Vision}, 113(1):54--66, 2015.

\bibitem{Chen2017Sep}
N.~F.~Y. Chen.
\newblock {Pseudo-labels for Supervised Learning on Dynamic Vision Sensor Data,
  Applied to Object Detection under Ego-motion}.
\newblock {\em arXiv}, Sep 2017.

\bibitem{Cohen2016Sep}
G.~K. Cohen.
\newblock {\em {Event-Based Feature Detection, Recognition and
  Classification}}.
\newblock PhD thesis, Universit{\ifmmode \acute{e} \else \'{e}\fi} Pierre et
  Marie Curie - Paris VI, Sep 2016.

\bibitem{dieh2015spiking}
P.~U. {Diehl}, D.~{Neil}, J.~{Binas}, M.~{Cook}, S.~{Liu}, and M.~{Pfeiffer}.
\newblock Fast-classifying, high-accuracy spiking deep networks through weight
  and threshold balancing.
\newblock In {\em 2015 International Joint Conference on Neural Networks
  (IJCNN)}, pages 1--8, July 2015.

\bibitem{Everingham2010Jun}
M.~Everingham, L.~Van~Gool, C.~K.~I. Williams, J.~Winn, and A.~Zisserman.
\newblock {The Pascal Visual Object Classes (VOC) Challenge}.
\newblock {\em Int. J. Comput. Vision}, 88(2):303--338, Jun 2010.

\bibitem{Fei-Fei2006Apr}
L.~Fei-Fei, R.~Fergus, and P.~Perona.
\newblock {One-shot learning of object categories}.
\newblock {\em IEEE Trans. Pattern Anal. Mach. Intell.}, 28(4):594--611, Apr
  2006.

\bibitem{gehrig2018asynchronous}
D.~Gehrig, H.~Rebecq, G.~Gallego, and D.~Scaramuzza.
\newblock Asynchronous, photometric feature tracking using events and frames.
\newblock In {\em Eur. Conf. Comput. Vis.(ECCV)}, 2018.

\bibitem{Glorot2010Mar}
X.~Glorot and Y.~Bengio.
\newblock {Understanding the difficulty of training deep feedforward neural
  networks}.
\newblock {\em PMLR}, pages 249--256, Mar 2010.

\bibitem{He2015Dec}
K.~He, X.~Zhang, S.~Ren, and J.~Sun.
\newblock Deep residual learning for image recognition.
\newblock In {\em Proceedings of the IEEE conference on computer vision and
  pattern recognition}, pages 770--778, 2016.

\bibitem{kim2019spiking}
S.~Kim, S.~Park, B.~Na, and S.~Yoon.
\newblock Spiking-yolo: Spiking neural network for real-time object detection.
\newblock {\em arXiv preprint arXiv:1903.06530}, 2019.

\bibitem{Kingma2014Dec}
D.~P. Kingma and J.~Ba.
\newblock {Adam: A Method for Stochastic Optimization}.
\newblock {\em arXiv}, Dec 2014.

\bibitem{kipf2017semi}
T.~N. Kipf and M.~Welling.
\newblock Semi-supervised classification with graph convolutional networks.
\newblock In {\em International Conference on Learning Representations (ICLR)},
  2017.

\bibitem{Krizhevsky2012}
A.~Krizhevsky.
\newblock Learning multiple layers of features from tiny images.
\newblock 04 2009.

\bibitem{Krizhevsky2012Dec}
A.~Krizhevsky, I.~Sutskever, and G.~E. Hinton.
\newblock Imagenet classification with deep convolutional neural networks.
\newblock In F.~Pereira, C.~J.~C. Burges, L.~Bottou, and K.~Q. Weinberger,
  editors, {\em Advances in Neural Information Processing Systems 25}, pages
  1097--1105. Curran Associates, Inc., 2012.

\bibitem{Lagorce2016Jul}
X.~Lagorce, G.~Orchard, F.~Galluppi, B.~E. Shi, and R.~B. Benosman.
\newblock {HOTS: A Hierarchy of Event-Based Time-Surfaces for Pattern
  Recognition}.
\newblock {\em IEEE Trans. Pattern Anal. Mach. Intell.}, 39(7), Jul 2016.

\bibitem{lagorce2017hots}
X.~Lagorce, G.~Orchard, F.~Galluppi, B.~E. Shi, and R.~B. Benosman.
\newblock Hots: a hierarchy of event-based time-surfaces for pattern
  recognition.
\newblock {\em IEEE transactions on pattern analysis and machine intelligence},
  39(7):1346--1359, 2017.

\bibitem{Lecun1998Nov}
Y.~Lecun, L.~Bottou, Y.~Bengio, and P.~Haffner.
\newblock {Gradient-based learning applied to document recognition}.
\newblock {\em Proc. IEEE}, 86(11):2278--2324, Nov 1998.

\bibitem{Li2017May}
H.~Li, H.~Liu, X.~Ji, G.~Li, and L.~Shi.
\newblock {CIFAR10-DVS: An Event-Stream Dataset for Object Classification}.
\newblock {\em Front. Neurosci.}, 11:309, May 2017.

\bibitem{Li2017}
J.~Li, F.~Shi, W.~Liu, D.~Zou, Q.~Wang, H.~Lee, P.-K. Park, and H.~E. Ryu.
\newblock {Adaptive temporal pooling for object detection using dynamic vision
  sensor}.
\newblock {\em British Machine Vision Conference (BMVC)}, 2017.

\bibitem{liu2018adaptive}
M.~Liu and T.~Delbruck.
\newblock Adaptive time-slice block-matching optical flow algorithm for dynamic
  vision sensors.
\newblock Technical report, 2018.

\bibitem{liu2016ssd}
W.~Liu, D.~Anguelov, D.~Erhan, C.~Szegedy, S.~Reed, C.-Y. Fu, and A.~C. Berg.
\newblock Ssd: Single shot multibox detector.
\newblock In {\em European conference on computer vision}, pages 21--37.
  Springer, 2016.

\bibitem{Long2014Nov}
J.~Long, E.~Shelhamer, and T.~Darrell.
\newblock Fully convolutional networks for semantic segmentation.
\newblock In {\em Proceedings of the IEEE conference on computer vision and
  pattern recognition}, pages 3431--3440, 2015.

\bibitem{Maass1997networks}
W.~Maass.
\newblock Networks of spiking neurons: the third generation of neural network
  models.
\newblock {\em Neural networks}, 10(9):1659--1671, 1997.

\bibitem{Maqueda_2018_CVPR}
A.~I. Maqueda, A.~Loquercio, G.~Gallego, N.~Garc\'{i}a, and D.~Scaramuzza.
\newblock Event-based vision meets deep learning on steering prediction for
  self-driving cars.
\newblock In {\em The IEEE Conference on Computer Vision and Pattern
  Recognition (CVPR)}, June 2018.

\bibitem{mitrokhin2018event}
A.~Mitrokhin, C.~Fermuller, C.~Parameshwara, and Y.~Aloimonos.
\newblock Event-based moving object detection and tracking.
\newblock {\em arXiv preprint arXiv:1803.04523}, 2018.

\bibitem{Mnih2014Jun}
V.~Mnih, N.~Heess, A.~Graves, et~al.
\newblock Recurrent models of visual attention.
\newblock In {\em Advances in neural information processing systems}, pages
  2204--2212, 2014.

\bibitem{monti2017geometric}
F.~Monti, D.~Boscaini, J.~Masci, E.~Rodola, J.~Svoboda, and M.~M. Bronstein.
\newblock Geometric deep learning on graphs and manifolds using mixture model
  cnns.
\newblock In {\em Proceedings of the IEEE Conference on Computer Vision and
  Pattern Recognition}, pages 5115--5124, 2017.

\bibitem{Mueggler2016Oct}
E.~Mueggler, H.~Rebecq, G.~Gallego, T.~Delbruck, and D.~Scaramuzza.
\newblock {The Event-Camera Dataset and Simulator: Event-based Data for Pose
  Estimation, Visual Odometry, and SLAM}.
\newblock {\em arXiv}, Oct 2016.

\bibitem{Neil2016Oct}
D.~Neil, M.~Pfeiffer, and S.-C. Liu.
\newblock Phased lstm: Accelerating recurrent network training for long or
  event-based sequences.
\newblock In {\em Advances in Neural Information Processing Systems}, pages
  3882--3890, 2016.

\bibitem{Orchard2015Nov}
G.~Orchard, A.~Jayawant, G.~K. Cohen, and N.~Thakor.
\newblock {Converting Static Image Datasets to Spiking Neuromorphic Datasets
  Using Saccades}.
\newblock {\em Front. Neurosci.}, 9, Nov 2015.

\bibitem{Perez-Carrasco2013Nov}
J.~A. P{\ifmmode \acute{e} \else \'{e}\fi}rez-Carrasco, B.~Zhao, C.~Serrano,
  B.~Acha, T.~Serrano-Gotarredona, S.~Chen, and B.~Linares-Barranco.
\newblock {Mapping from frame-driven to frame-free event-driven vision systems
  by low-rate rate coding and coincidence processing--application to
  feedforward ConvNets}.
\newblock {\em IEEE Trans. Pattern Anal. Mach. Intell.}, 35(11):2706--2719, Nov
  2013.

\bibitem{Posch2011Jan}
C.~Posch, D.~Matolin, and R.~Wohlgenannt.
\newblock {A QVGA 143 dB Dynamic Range Frame-Free PWM Image Sensor With
  Lossless Pixel-Level Video Compression and Time-Domain CDS}.
\newblock {\em IEEE J. Solid-State Circuits}, 46(1):259--275, Jan 2011.

\bibitem{Raj2015}
A.~Raj, D.~Maturana, and S.~Scherer.
\newblock Multi-scale convolutional architecture for semantic segmentation.
\newblock page~14, 01 2015.

\bibitem{Ramesh2017Oct}
B.~Ramesh, H.~Yang, G.~Orchard, N.~A.~L. Thi, and C.~Xiang.
\newblock {DART: Distribution Aware Retinal Transform for Event-based Cameras}.
\newblock {\em arXiv}, Oct 2017.

\bibitem{rameshlong}
B.~Ramesh, S.~Zhang, Z.~W. Lee, Z.~Gao, G.~Orchard, and C.~Xiang.
\newblock Long-term object tracking with a moving event camera.
\newblock 2018.

\bibitem{rebecq2017evo}
H.~Rebecq, T.~Horstschaefer, G.~Gallego, and D.~Scaramuzza.
\newblock Evo: A geometric approach to event-based 6-dof parallel tracking and
  mapping in real time.
\newblock {\em IEEE Robotics and Automation Letters}, 2(2):593--600, 2017.

\bibitem{Redmon2015Jun}
J.~Redmon, S.~Divvala, R.~Girshick, and A.~Farhadi.
\newblock You only look once: Unified, real-time object detection.
\newblock In {\em Proceedings of the IEEE conference on computer vision and
  pattern recognition}, pages 779--788, 2016.

\bibitem{Ren2015Jun}
S.~Ren, K.~He, R.~Girshick, and J.~Sun.
\newblock Faster r-cnn: Towards real-time object detection with region proposal
  networks.
\newblock In {\em Advances in neural information processing systems}, pages
  91--99, 2015.

\bibitem{Serrano-Gotarredona2013Mar}
T.~Serrano-Gotarredona and B.~Linares-Barranco.
\newblock {A $128 \times 128$ 1.5 $\%$ Contrast Sensitivity 0.9 $\%$ FPN 3
  $\mu$s Latency 4 mW Asynchronous Frame-Free Dynamic Vision Sensor Using
  Transimpedance Preamplifiers}.
\newblock {\em IEEE J. Solid-State Circuits}, 48(3):827--838, Mar 2013.

\bibitem{Serrano-Gotarredona2015Dec}
T.~Serrano-Gotarredona and B.~Linares-Barranco.
\newblock {Poker-DVS and MNIST-DVS. Their History, How They Were Made, and
  Other Details}.
\newblock {\em Front. Neurosci.}, 9, Dec 2015.

\bibitem{Simonyan2014}
K.~Simonyan and A.~Zisserman.
\newblock Very deep convolutional networks for large-scale image recognition.
\newblock {\em CoRR}, abs/1409.1556, 2014.

\bibitem{sironi2018hats}
A.~Sironi, M.~Brambilla, N.~Bourdis, X.~Lagorce, and R.~Benosman.
\newblock Hats: Histograms of averaged time surfaces for robust event-based
  object classification.
\newblock In {\em Proceedings of the IEEE Conference on Computer Vision and
  Pattern Recognition}, pages 1731--1740, 2018.

\bibitem{stoffregen2018simultaneous}
T.~Stoffregen and L.~Kleeman.
\newblock Simultaneous optical flow and segmentation (sofas) using dynamic
  vision sensor.
\newblock {\em arXiv preprint arXiv:1805.12326}, 2018.

\bibitem{Szegedy2016Feb}
C.~Szegedy, S.~Ioffe, V.~Vanhoucke, and A.~A. Alemi.
\newblock Inception-v4, inception-resnet and the impact of residual connections
  on learning.
\newblock In {\em AAAI}, volume~4, page~12, 2017.

\bibitem{Yu2015Nov}
F.~Yu and V.~Koltun.
\newblock {Multi-Scale Context Aggregation by Dilated Convolutions}.
\newblock In {\em International Conference on Learning Representations (ICLR)},
  2016.

\end{thebibliography}


\begin{thebibliography}{1}\itemsep=-1pt

\bibitem{Blender2017}
{Blender Online Community}.
\newblock {\em Blender - a 3D modelling and rendering package}.
\newblock Blender Foundation, Blender Institute, Amsterdam, 2017.

\bibitem{Ester1996}
M.~Ester, H.-P. Kriegel, J.~Sander, and X.~Xu.
\newblock A density-based algorithm for discovering clusters a density-based
  algorithm for discovering clusters in large spatial databases with noise.
\newblock In {\em Proceedings of the Second International Conference on
  Knowledge Discovery and Data Mining}, KDD'96, pages 226--231. AAAI Press,
  1996.

\bibitem{Lecun1998Nov}
Y.~Lecun, L.~Bottou, Y.~Bengio, and P.~Haffner.
\newblock {Gradient-based learning applied to document recognition}.
\newblock {\em Proc. IEEE}, 86(11):2278--2324, Nov 1998.

\bibitem{Mueggler2016Oct}
E.~Mueggler, H.~Rebecq, G.~Gallego, T.~Delbruck, and D.~Scaramuzza.
\newblock {The Event-Camera Dataset and Simulator: Event-based Data for Pose
  Estimation, Visual Odometry, and SLAM}.
\newblock {\em arXiv}, Oct 2016.

\bibitem{Orchard2015Nov}
G.~Orchard, A.~Jayawant, G.~K. Cohen, and N.~Thakor.
\newblock {Converting Static Image Datasets to Spiking Neuromorphic Datasets
  Using Saccades}.
\newblock {\em Front. Neurosci.}, 9, Nov 2015.

\bibitem{Serrano-Gotarredona2015Dec}
T.~Serrano-Gotarredona and B.~Linares-Barranco.
\newblock {Poker-DVS and MNIST-DVS. Their History, How They Were Made, and
  Other Details}.
\newblock {\em Front. Neurosci.}, 9, Dec 2015.

\bibitem{Stromatias2017Jun}
E.~Stromatias, M.~Soto, T.~Serrano-Gotarredona, and B.~Linares-Barranco.
\newblock {An Event-Driven Classifier for Spiking Neural Networks Fed with
  Synthetic or Dynamic Vision Sensor Data}.
\newblock {\em Front. Neurosci.}, 11, Jun 2017.

\end{thebibliography}
\end{document}